\documentclass[10pt,journal,compsoc]{IEEEtran}
\usepackage[nocompress]{cite}
\usepackage[pdftex]{graphicx}
\graphicspath{{./figs/}}
\usepackage{times}
\usepackage{epsfig}
\usepackage{graphicx}
\usepackage{amsmath}
\usepackage{amssymb}
\usepackage{amsthm}
\usepackage{booktabs}
\usepackage{subfigure}
\usepackage{algorithm}
\usepackage{algorithmicx}  
\usepackage{algpseudocode} 
\usepackage{arydshln}
\usepackage{multirow}
\usepackage{footnote}
\usepackage{threeparttable}
\usepackage[usenames,dvipsnames]{color}

\usepackage{silence}
\def\mbfx{\mathbf{x}}
\def\mbfz{\mathbf{z}}

\def\mbfh{\mathbf{h}}
\def\mbfv{\mathbf{v}}
\def\mbfu{\mathbf{u}}

\def\mbfeps{\boldsymbol{\varepsilon}}

\def\mbfeps{\boldsymbol{\varepsilon}}

\newcommand{\ie}{{\emph{i.e.}}}

\newcommand{\eg}{{\emph{e.g.}}}
\newcommand{\etc}{etc.}
\newcommand{\etal}{{\emph{et al.}}}
\def\degree{^{\circ}}
\def\sArt{state-of-the-art~}

\newcommand{\figref}[1]{Fig.~\ref{#1}}

\newcommand{\refeq}[1]{Eq.~(\ref{eq:#1})}
\newtheorem{theorem}{Theorem}

\newcommand{\todo}[1]{#1}

\begin{document}

\title{Class-Specific Semantic Reconstruction for Open Set Recognition}

\author{Hongzhi~Huang, Yu~Wang, Qinghua~Hu~\IEEEmembership{Senior Member,~IEEE,}
        Ming-Ming~Cheng~\IEEEmembership{Senior Member,~IEEE}
\IEEEcompsocitemizethanks{
\IEEEcompsocthanksitem H. Huang, Y. Wang, and Q. Hu are with the College of Intelligence and Computing, Tianjin University, Tianjin 300350, China. Y. Wang, and Q. Hu are also with Haihe Laboratory of Information Technology Application Innovation, Tianjin, China.
M.M. Cheng is with the TKLNDST, College of Computer Science, Nankai University, 
Tianjin 300350, China.
\IEEEcompsocthanksitem Y. Wang and Q. Hu are the corresponding authors of 
this paper (\{wangyu\_, huqinghua\}@tju.edu.cn).
\IEEEcompsocthanksitem This work was supported in part by the National Key Research and Development Program of China under Grant 2019YFB2101901, in part by the National Natural Science Foundation of China under Grants 62106174, 61732011, and 61876127, and in part by the China Postdoctoral Science Foundation under Grants 2021TQ0242 and 2021M690118.
}
}

\markboth{IEEE TRANSACTIONS ON PATTERN ANALYSIS AND MACHINE INTELLIGENCE}
{Shell \MakeLowercase{\textit{et al.}}: Bare Demo of IEEEtran.cls for Computer Society Journals}

\IEEEtitleabstractindextext{%
\begin{abstract}
    Open set recognition enables deep neural networks (DNNs) to identify samples of unknown classes, while maintaining high classification accuracy on samples of known classes. Existing methods basing on auto-encoder (AE) and prototype learning show great potential in handling this challenging task. In this study, we propose a novel method, called Class-Specific Semantic Reconstruction (CSSR), that integrates the power of AE and prototype learning.
   Specifically, CSSR replaces prototype points with manifolds represented by class-specific AEs. Unlike conventional prototype-based methods, CSSR models each known class on an individual AE manifold, and measures class belongingness through AE's reconstruction error. Class-specific AEs are plugged into the top of the DNN backbone and reconstruct the semantic representations learned by the DNN instead of the raw image. Through end-to-end learning, the DNN and the AEs boost each other to learn both discriminative and representative information. The results of experiments conducted on multiple datasets show that the proposed method achieves outstanding performance in both close and open set recognition and is sufficiently simple and flexible to incorporate into existing frameworks.
\end{abstract}

}

\maketitle

\IEEEdisplaynontitleabstractindextext

\IEEEpeerreviewmaketitle

\IEEEraisesectionheading{\section{Introduction}\label{sec:introduction}}



\IEEEPARstart{C}ONVENTIONAL deep neural networks (DNNs) 
are trained based on a closed-set assumption, 
in which the test classes are all seen during training. 
In real-world applications, the test samples may come from unknown classes
\cite{yoshihashi2019classification}.  
When meeting such an unknown sample, 
traditional DNNs will compulsorily classify it as one of the known classes 
and make a wrong prediction, 
which may lead to irreparable losses in certain critical scenarios, 
such as medical diagnosis and autonomous driving.


Open set recognition (OSR) addresses this challenge by making the models correctly 
classify the samples from known classes (\ie, the closed set) 
and accurately identify those from unknown classes
(\ie, the open set) \cite{geng2020recent}. 
The main challenge for OSR is that no information about unknown classes 
is available during training, 
making it difficult to distinguish known and unknown classes
(\ie, reduce open space risk) \cite{scheirer2013toward}. 
%
Traditional DNNs emphasize the discriminative features 
of known classes, and learn a partition of the entire feature space. 
%
This leads to a serious problem: samples of unknown classes are still located in certain specific regions and, thus, are identified as known classes with high confidence \cite{prototype}. 
Consequently, many previous works have proposed to learn compact 
representations of known classes, 
so that the model can separate closed and open set spaces. 
Among these methods, auto-encoder (AE)-based methods \cite{oza2019c2ae,
oza2019deep,sun2020conditional,yoshihashi2019classification,sun2020open} 
and prototype-like methods 
\cite{chen2020learning,prototype,chen2021adversarial} 
are currently the most powerful.

\begin{figure*}[tb]
  \centering
  \includegraphics[width=\linewidth]{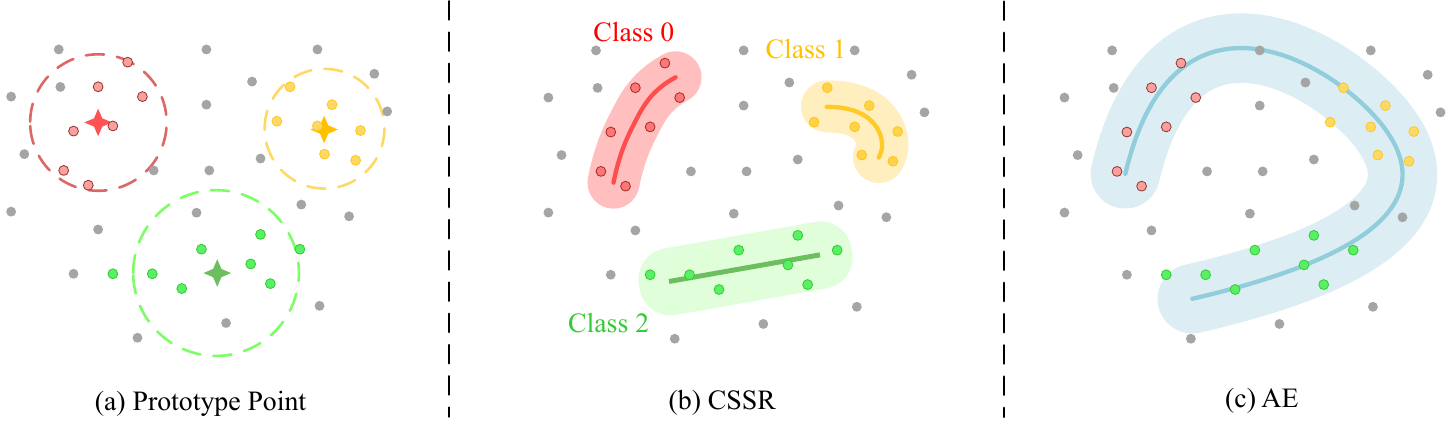}
  \caption{Comparison of open set methods using prototype learning (a), proposed method (b), and auto-encoder (c). 
    Prototype learning models each class with several prototype points, 
    and it has limited fitting ability and 
    learns completely compact representations. 
    Auto-encoder models known samples with a continuous 
    low-dimensional manifold but some inter-class regions are 
    devoured and misclassified into a known space. 
    CSSR models each known class on an individual manifold 
    represented by an AE. It fits known samples well while releasing the regions devoured by a single AE.
  }\label{fig:overview}
\end{figure*}

As illustrated in \figref{fig:overview}(c),
AE-based methods learn latent representations by reconstructing 
the raw input to retain most information from an image. 
As an AE can learn to reconstruct the images of known classes during training, 
test images originating from unknown classes would lead to high reconstruction 
errors, and thus, can be recognized \cite{oza2019c2ae,sun2020conditional}. 
This can be considered learning a low-dimensional manifold to fit the 
distribution of known samples.
To classify known classes, 
these methods learn a classifier based on latent representations 
obtained through pixel-wise reconstruction of the raw image. 
However, two problems remain in these methods: (1) classification degradation and
(2) open space risk introduction. 

In classification degradation, using the latent representations learned by AE for the classifier harms the performance of closed-set classification. This occurs primarily because some unnecessary classification information (e.g., background information) is retained and interferes with the learning of the classifier on recognizing known classes~\cite{tian2020what}. Open space risk introduction refers to the fact that continuous manifold learned to fit known samples might devour the inter-class regions and, thus, introduce open space risk (\figref{fig:overview}(c)).

Unlike AE-based methods, prototype-like methods, 
including generalized convolutional prototype learning \cite{prototype} 
and the recently proposed reciprocal point learning \cite{chen2020learning}, 
learn class-specific points to fit the extracted representations 
corresponding to the labeled class or rest classes, 
respectively. 
The sense of these methods is straightforward. However, the prototype-learning framework still faces great challenges on the OSR task. The main challenge is the \textbf{class under-representation problem}, where using only a single point or very few points cannot sufficiently represent the class. On the one hand, prototype learning assumes the Gaussian distribution for class-specific features \cite{prototype}. However, this is seldom satisfied in real-world applications, which would introduce the open space risk. On the other hand, intra-class features are compressed to a quite limited number of points in prototype-learning frameworks. This may cause the model to filter out certain necessary information that could help discriminate unknown classes \cite{chu2020distance}.

%

To address the aforementioned problems of both AE-based and prototype-like learning methods, 
we fully exploit them in a proposed novel method, 
called class-specific semantic reconstruction (CSSR).
Specifically, CSSR reduces open space risk by modeling each known class with a specific 
AE manifold. 
In CSSR, the feature of each sample is extracted using a DNN.
Then, an individual AE is specified to each known class to project 
different classes to various manifolds. 
The AEs are plugged into the top of the DNN backbone to reconstruct 
the semantic representations, rather than the raw image. 
As the AE manifold is a learnable representation of specific categories, 
the reconstruction errors, signifying the point-to-manifold distance, 
are used as the logits for classification. 
%
CSSR minimizes the cross-entropy loss, where the reconstruction error 
of the AE specified to the labeled category is minimized. 
A graphical description is shown in \figref{fig:overview}b. 

Our proposed framework can help solve the aforementioned problems of existing methods.
Compared to AE-based methods, the proposed CSSR approach (1) solves the problem of classification degradation by discarding unnecessary information and reconstructing the semantic features rather than the raw image and (2) handles the open space risk problem by learning the class-specific manifolds to release the devoured 
inter-class regions. Compared to prototype-like methods, the proposed CSSR approach deals well with the problem of class under-representation by learning a class-specific manifold. This not only breaks the Gaussian assumption of classes but also retains more key information of classes than representing the class with a single point. 
%
%

Through the end-to-end learning process, 
the class-specific AEs and DNN boost each other to identify the open space
while learning highly class-related semantic representations. 
The AEs tend to associate each class with a subset of semantic features. 
The sample of a known class tends to activate its related features
while inactivating unrelated ones. 
For samples of an unknown class, 
their semantic features are not activated as they are not related to 
any features of known classes. 
This property is also exploited for detecting unknown classes. 
The results of the experiments conducted on various datasets show that 
the proposed method significantly outperforms other \sArt methods,
and improves the performance of both closed and open set recognition. 

In summary, this study makes the following contributions:
\begin{enumerate}
  \item We propose a simple yet effective method, namely CSSR, for open set recognition. It specifies an individual AE to each known class, and plugs such AEs into the top of the DNN  backbone to reconstruct the semantic representations learned by the backbone network. CSSR improves the fitting and representation learning ability, thereby enhancing the open set performance.
  \item We conduct a theoretical analysis to explain the open space risk produced by existing methods, and discuss the connections between CSSR and existing methods to understand CSSR comprehensively.
  \item We performed experiments under various protocols. The results demonstrate that CSSR can significantly outperform baseline methods and achieve state-of-the-art performance on multiple public datasets. Typically, CSSR improves F1 score by an average of 8.3\% on the task of open set recognition. 
\end{enumerate}

\section{Related Work}

Our work is mainly related to open set recognition, particularly AE-based and prototype-like methods. Open set recognition is naturally related to some other problems (\eg, out-of-distribution (OOD) detection \cite{hendrycks2016a} and novelty detection \cite{perera2019ocgan}). The OOD detection methods are briefly discussed in this section.

\subsection{Open Set Recognition}

Early works utilized traditional machine learning methods. They used the scores produced by the classifiers, and the unknown samples could be identified by measuring the similarity between the samples and known classes \cite{bendale2015towards, junior2017nearest}. For example,
Scheirer \etal \cite{scheirer2013toward} employed a support vector machine for known class identification and adopted the extreme value distribution to detect the unknowns. Recently, the powerful representation learning capability of DNNs has been applied to detect the unknowns. 

Several scholars have designed or utilized the classification layer for OSR \cite{hendrycks2016a,bendale2016towards,zhou2021learning}. A plain choice is to utilize the maximum SoftMax probabilities and reject unconfident predictions \cite{hendrycks2016a}. Bendale \etal \cite{bendale2016towards} proved SoftMax probability is not robust and proposed to replace the SoftMax function with the OpenMax function, which redistributes the scores of SoftMax to obtain the confident score of the unknown class explicitly. 
Zhou \etal \cite{zhou2021learning} proposed the concept of placeholder learning, where the overconfident predictions are calibrated by reserving classifier placeholders for unknown classes.

\textbf{AE-based DNN Methods}.
Zhang \etal \cite{zhang2017sparse} asserted that reconstruction errors contain useful discriminative information, and proposed to use sparse representation to model the open set recognition problem.
Yoshihashi \etal \cite{yoshihashi2019classification} designed the CROSR method, which uses latent representations for closed set classifier training and unknown detection. Oza and Patel \cite{oza2019c2ae} proposed a two-step C2AE method. The method first trains the encoder for closed set identification, then keeps it fixed and adds the class-conditional information to train the decoder for unknown detection. Sun \etal \cite{sun2020conditional} used a variational AE to force different latent features to approximate different Gaussian models for unknown detection. They subsequently developed CPGM \cite{sun2020open}, which adds discriminative information into probabilistic generative models. Perera \etal \cite{perera2020generative} fed the raw and reconstructed images to a classification network, and the prediction was simultaneously confident when the reconstruction was consistent with the raw input. However, AE-based methods suffer from two problems, as stated in Section \ref{sec:introduction}. (1) The representation learned from pixelwise image reconstruction contains unnecessary background information. This might harm both close and open set performance. (2) AEs learn a continuous manifold to fit known samples, which might devour the inter-class regions.

\textbf{Prototype-like DNN Methods}. Yang \etal \cite{prototype} proposed generalized convolutional prototype learning, which replaces the close-world assumed SoftMax classifier with an open-world oriented prototype model. Chen \etal \cite{chen2020learning} developed reciprocal point learning (RPL), which classifies a sample as known or unknown based on the otherness with reciprocal points. Subsequently, RPL was further improved to ARPL \cite{chen2021adversarial}, integrating an extra adversarial training strategy to enhance the model distinguishability into the known and unknown classes by generating confusing training samples. Prototype-like methods are limited owing to the lack of fitting ability and representation diversity. We apply class-specific AEs to address this issue.

\subsection{OOD Detection}
As first introduced by Hendrycks and Gimpel \cite{hendrycks2016a}, OOD detection involves the detection of samples that do not belong to the training set. Several methods have considered the problem where OOD samples are available during training \cite{lee2018a,liang2018enhancing,dhamija2018reducing,liu2020energy,quintanilha2018detecting,zisselman2020deep}. However, this is not congruent with our task, where only in-distribution data are accessible during training. \todo{Though OOD detection can be greatly simplified with available known outliers, open set recognition is more common and realistic in practical. Moreover, well-designed models can also be benefited from outlier exposure, \eg, OpenGAN~\cite{kong2021opengan} can work with or without auxiliary OOD data.}
In the following, we mainly focus on the models trained without extra OOD data.

\textbf{Supervised Methods.} With a similar problem setting to open set recognition, these methods build OOD detectors upon a classification task. Some methods seek better score functions, including maximum SoftMax probability \cite{hendrycks2016a}, maximum logit scores \cite{hendrycks2019a}, and energy score \cite{liu2020energy}. Vyas \etal \cite{vyas2018out} used an ensemble of leave-one-out classifiers to simulate OOD accessible training individually. Sastry and Oore \cite{sastry2020detecting} proposed to characterize activity patterns with Gram matrices and score OOD-ness by calculating the element-wise deviation comparing the Gram matrices from the training data.

\textbf{Self-Supervised Methods.} These methods are relatively novel and exploit the well-learned representations by self-supervision. Golan and El-Yaniv \cite{golan2018deep}, and Hendrycks \etal \cite{hendrycks2019using} considered the task of predicting image transformations (\eg, rotating image to $0\degree,90\degree,180\degree, \text{and}\quad 270\degree$), which has also been leveraged as an auxiliary task in \cite{tack2020csi}. Self-supervised contrastive learning has shown considerable success in unsupervised representation learning \cite{chen2020a,grill2020bootstrap}, and is being applied to OOD detection \cite{tack2020csi,winkens2020contrastive,sehwag2021ssd}. They observed that representations obtained by contrastive learning have distinct patterns between in- and out-distribution data. Although designed for unlabeled settings, these methods have also been extended for supervised learning.
\section{Preliminaries}

\textbf{Open Set Recognition.} Given a set of $n$ labeled instances, $\mathcal{X} = \{(\mbfx_i,y_i)\}_{i=1}^n$, where $y_i\in \{1,...,m\}$ are the corresponding labels of known classes. For open set recognition, the goal is to learn a model from $\mathcal{X}$ that classifies test samples into $m+1$ classes, \ie, one of the $m$ known classes or an unknown class indexed by $m+1$.

\textbf{Auto-Encoder.} An AE learns effective representations of a set of data in an unsupervised manner. With the bottleneck structure of cascaded encoder $f$ and decoder $g$, AE is forced to compress high-dimensional input features to a low-dimension embedding space $H$ to adequately reconstruct the raw input, \ie, minimizing the reconstruction error $\|\mbfx - g(f(\mbfx))\|_2^2$ for each input sample $\mbfx$. The decoder $g$ learns a manifold $V=\{g(\mbfh)|\mbfh \in H\}$, while the encoder $f$ learns a mapping from the original feature space to the manifold $V$. AE-based open set recognition methods fit manifold $V$ to the distribution of known class samples, where the reconstruction error is the distance metric between the input sample and manifold $V$. Existing methods reconstruct the entire image and minimize pixel-wise reconstruction error. However, fitting background pixels (category irrelevant information) is unhelpful for both close set and open set recognition; this has also been demonstrated by Zhang \etal \cite{zhang2020hybrid}, who reported that building a flow density estimator on latent representation works better than on the raw image. Therefore, we build AEs on the latent space extracted by a backbone network.  

\textbf{Prototype and Reciprocal Learning.} By defining class-specific points set $U_i$ for each category $i$, prototype learning \cite{prototype} assigns a test sample to the nearest prototype point, and samples that are far away from all prototype points are regarded as being from an unknown class. Formally, it models closed and open set recognition as follows:
\begin{align}
    p(y=i|\mbfx, \mathcal{B}, U) \propto& \left(-\min_{\mbfu \in U_i}{\|\mathcal{B}(\mbfx) - \mbfu\|^2_2} \right), \notag \\
    p(unknown|\mbfx, \mathcal{B}, U) \propto& \min_i \min_{\mbfu \in U_i}{\|\mathcal{B}(\mbfx) - \mbfu\|^2_2}, \label{eq:prototype}
\end{align}
where $\mathcal{B}$ is the backbone network extracting the embedding feature from input $\mbfx$.
Contrarily, reciprocal learning \cite{chen2020learning} utilizes class-specific reciprocal point set $U_i$ to learn otherness, instead of belongingness, and considers samples close to all reciprocal points to be unknown, which is expressed as
\begin{align}
    p(y=i|\mbfx, \mathcal{B}, U) \propto& \sum_{\mbfu \in U_i}{\|\mathcal{B}(\mbfx) - \mbfu\|^2_2}, \notag \\
    p(known|\mbfx, \mathcal{B}, U) \propto& \max_i \max_{\mbfu \in U_i}{\|\mathcal{B}(\mbfx) - \mbfu\|^2_2}. \label{eq:reciprocal}
\end{align}
In the training phase, the model optimizes the cross-entropy loss on SoftMax normalized class probabilities. However, optimizing discriminative loss alone is ineffective. Therefore, both methods propose different regularization terms to manage open space risk and achieve better training. The prototype framework proposes a generative loss $\mathcal{L}_{pl}$ (also called prototype loss), which is the maximum likelihood regularization under the Gaussian mixture density assumption:
\begin{align}
    \mathcal{L}_{pl}(\mbfx,y;U,\mathcal{B})= \min_{\mbfu \in U_y} \|\mathcal{B}(\mbfx) - \mbfu\|_2^2. \label{eq:prototypeloss}
\end{align}
For the reciprocal learning framework, open space risk is bounded by the constraining variance of feature-to-reciprocal point distances. This is formalized by
\begin{align}
    \mathcal{L}_{rp}(\mbfx,y;U,\mathcal{B})=\sum_{u\in U_y} \|d(\mathcal{B}(\mbfx),u)-R_y\|_2^2, \label{eq:rp_reg}
\end{align}
where $d(\cdot,\cdot)$ is a distance function and $R_y$ is a class-specific learnable margin. 

Both $\mathcal{L}_{pl}$ and $\mathcal{L}_{rp}$ introduce extra compactness constraints. In this study, we observe that optimizing single discriminative cross entropy loss leads to inconsistent distribution between feature distribution and prototype points. Refer to Section \ref{sec:fitting_known} for a detailed discussion.
\section{Method}

As has been reported, prototype learning suffers from two problems: (1) class-specific prototype point sets have limited fitting ability; (2) non-diverse representation is insufficient for OSR. To improve the fitting ability and representation learning ability in the prototype-learning framework, we propose to leverage the power of AEs, which generates prototype manifolds to fit known classes. In this section, we first introduce our major architecture, and analyze how the proposed model manages open space risk. Then, we describe the proposed unknown detection strategy. Finally, the connections between our model and existing methods are discussed.

\subsection{Class-Specific Semantic Reconstruction}

\begin{figure}
\begin{center}
    \includegraphics[width=0.95\linewidth]{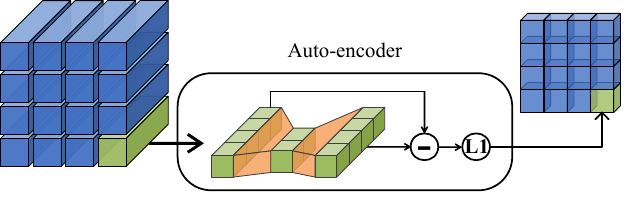}
    \caption{ Structure of individual AEs. Note that we operate the reconstruction process at the pixel level. It takes the semantic feature map as input and outputs the reconstruction error at each pixel. 
    }
    \label{fig:autoencoder}
\end{center}
\end{figure}

We replace class-specific point set $U_i$ with an AE for each category $i$ denoted as $\mathcal{A}_i$. As shown in \figref{fig:autoencoder}, it takes a latent representation $\mbfz$ as input, and outputs the reconstructed representation $\hat{\mbfz}=\mathcal{A}_i(\mbfz)$. Then, we calculate the reconstruction error by the L1-norm as
\begin{align}
    d(\mbfz,\mathcal{A}_i) = \|\mbfz - \mathcal{A}_i(\mbfz)\|_1. \label{eq:rcerror}
\end{align}
Following prototype learning, based on the reconstruction error, our framework can estimate class belongingness. Given sample $(\mbfx,c) \in \mathcal{X}$, we let $p(y = i | \mbfx) \propto (-d(\mbfz,\mathcal{A}_i))$ to learn the prototype manifold. 
By applying SoftMax to normalize the logits, the final probability can be defined as
\begin{align}
    p(y=i|\mbfz,\mathcal{A})=\frac{e^{-\gamma d(\mbfz,\mathcal{A}_i)}}{\sum_{j=1}^m{e^{-\gamma d(\mbfz,\mathcal{A}_j)}}},
\end{align}
where $\gamma$ is a hyper-parameter that controls the hardness of probability assignment. 
\todo{Considering an ideal solution to maximize output probability for the ground-truth category, the AEs should first learn a min-distance mapping (from feature to manifold) to minimize reconstruction error for the ground-truth category. Meanwhile, the manifolds of AEs should also learn to keep distance from each other to maximize the reconstruction error for the AEs except the one corresponds to the ground truth. }

As described in the preliminaries, class-specific AE $\mathcal{A}_i$ defines class-specific manifold $V_i$.
\todo{Under the above ideal circumstances, maximizing $p(y=c|\mbfx,\mathcal{A})$ can be considered prototype learning with an infinite number of prototype points (manifold $V_i$).} Assuming $d(\mbfz,\mathcal{A}_c)$ is minimized, the reconstruction error can be approximately expressed as
\begin{align}
    d(\mbfz,\mathcal{A}_c)= \|\mbfz - \mathcal{A}_c(\mbfz)\|_1 \approx \min_{\mbfv\in V_c} \|\mbfz - \mbfv \|_1. \label{eq:approx_prototype}
\end{align}
This process of point assignment is equivalent to (\ref{eq:prototype}), 
with point set $U_c$ replaced by $V_c$ and the squared L2-norm replaced by the L1-norm.

\subsection{Managing Open Space Risk}
\label{sec:openspacerisk}

\begin{figure}
\begin{center}
    \subfigure[CSSR]{\label{fig:openspace2} \includegraphics[width=0.48\linewidth]{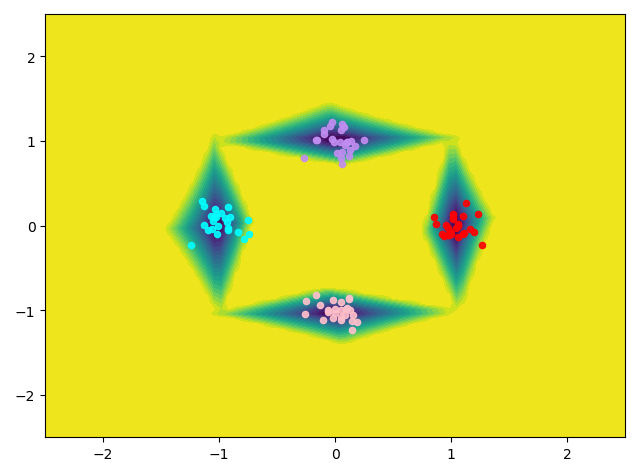}}
    \subfigure[CSSR MSE]{\label{fig:openspace3} \includegraphics[width=0.48\linewidth]{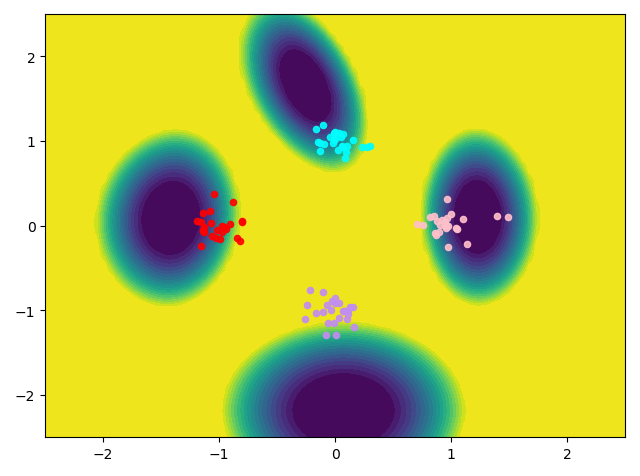}}
    \caption{Comparison of open space modeled by (a) CSSR, and (b) CSSR with mean squared reconstruction error. We set up a trivial experiment, where four classes generated from different Gaussian distributions are used for training. Then, we tested the entire feature space for unknown detection. The bright yellow and dark blue regions correspond to the open space and the close space identified by the different methods, respectively.}
    \label{fig:openspace}
\end{center}
\end{figure}

\subsubsection{Fitting Known Classes}
\label{sec:fitting_known}

In prototype point learning, mean square error (MSE) is used as the distance measure. We observe that inconsistent distribution between prototype points and semantic features is likely to be learned when using MSE. The example in \figref{fig:openspace3} visualizes such phenomenon, where a gap is shown between the prototypes and ground truth distribution. We also observe that such a gap results from the model's over-fitting, rather than under-fitting. The prototype loss given by \refeq{prototypeloss}, as a regularization term in GCPL, is attempting to explicitly reduce such a gap, and thus, manage open space risk. 
We next provide an analysis on how MSE leads to the aforementioned inconsistency, whereas mean absolute error, as we used in \refeq{rcerror}, keeps the consistency.

We consider the simplest form of prototype learning in the following analysis, where $|U_i| = 1$ and $\mbfu_i \in U_i$ represents the only prototype point for class $i$. Therefore, $d(\mbfz,U_i)$ is simplified to $\| \mbfz - \mbfu_i \|$, and the SoftMax probability for category $i$ is
\begin{align}
    p(y=i|\mbfz,U) = \frac{\exp(-\| \mbfz - \mbfu_i \|)}{\sum_j \exp (-\| \mbfz - \mbfu_j \| ) }. \label{eq:pprob}
\end{align}
Considering $\mbfz = \mbfu_c + \mbfeps$, we analyze how $p(y=c|\mbfz,U)$ varies from $\mbfeps$ with MAE and MSE used, respectively. \todo{The purpose of prototype learning is to let $p(y=c|\mbfz,U)$ be maximized if a sample locates exactly at its prototype point, \ie, $\mbfz = \mbfu_c$. The probability should decrease as the offset $\mbfeps = \mbfz - \mbfu_c$ gets large. However, the following theorems show that the above purpose cannot be achieved using MSE.}
\begin{theorem}
\label{thr1}
\todo{With $d(\mbfz,U_i) = \| \mbfz - \mbfu_i \|_2^2$, assuming $\ \mbfu_i \neq \mbfu_j$ for $\forall i\neq j$, there exists $c$ and $\mbfeps\neq 0$ satisfying $p(y=c|\mbfu_c,U) < p(y=c|\mbfu_c + \mbfeps,U)$.}
\end{theorem}

\begin{theorem}
\label{thr2}
With $d(\mbfz,U_i) = \| \mbfz - \mbfu_i \|_1$,
for each $c,\mbfeps$, $p(y=c|\mbfu_c,U) \ge p(y=c|\mbfu_c + \mbfeps,U)$ stands.
\end{theorem}



The proof of the two theorems can be found in the supplemental file. Let $c$ be the label with respect to $\mbfz$;
then, Theorem \ref{thr1} indicates that, with MSE used, $\mbfz = \mbfu_c$ may not be the best solution
to maximize $p(y=c|\mbfz ,U)$, and, to minimize the cross-entropy loss, thereby causing the aforementioned inconsistent distribution.
In contrast, MAE strictly satisfies $p(y=c|\mbfu_c,U) \ge p(y=c|\mbfu_c + \mbfeps,U)$, as Theorem \ref{thr2} suggests;
hence the maximum $p(y=c|\mbfz ,U)$ and minimum cross entropy loss is taken when $\mbfz = \mbfu_c$,
which guarantees consistency between distributions of the semantic feature and prototype point.
The experiment illustrated in \figref{fig:openspace2} shows that the prototypes are well fitted, and that the open space risk is well managed.


\begin{figure}
\begin{center}
    \subfigure{\includegraphics[width=0.48\linewidth]{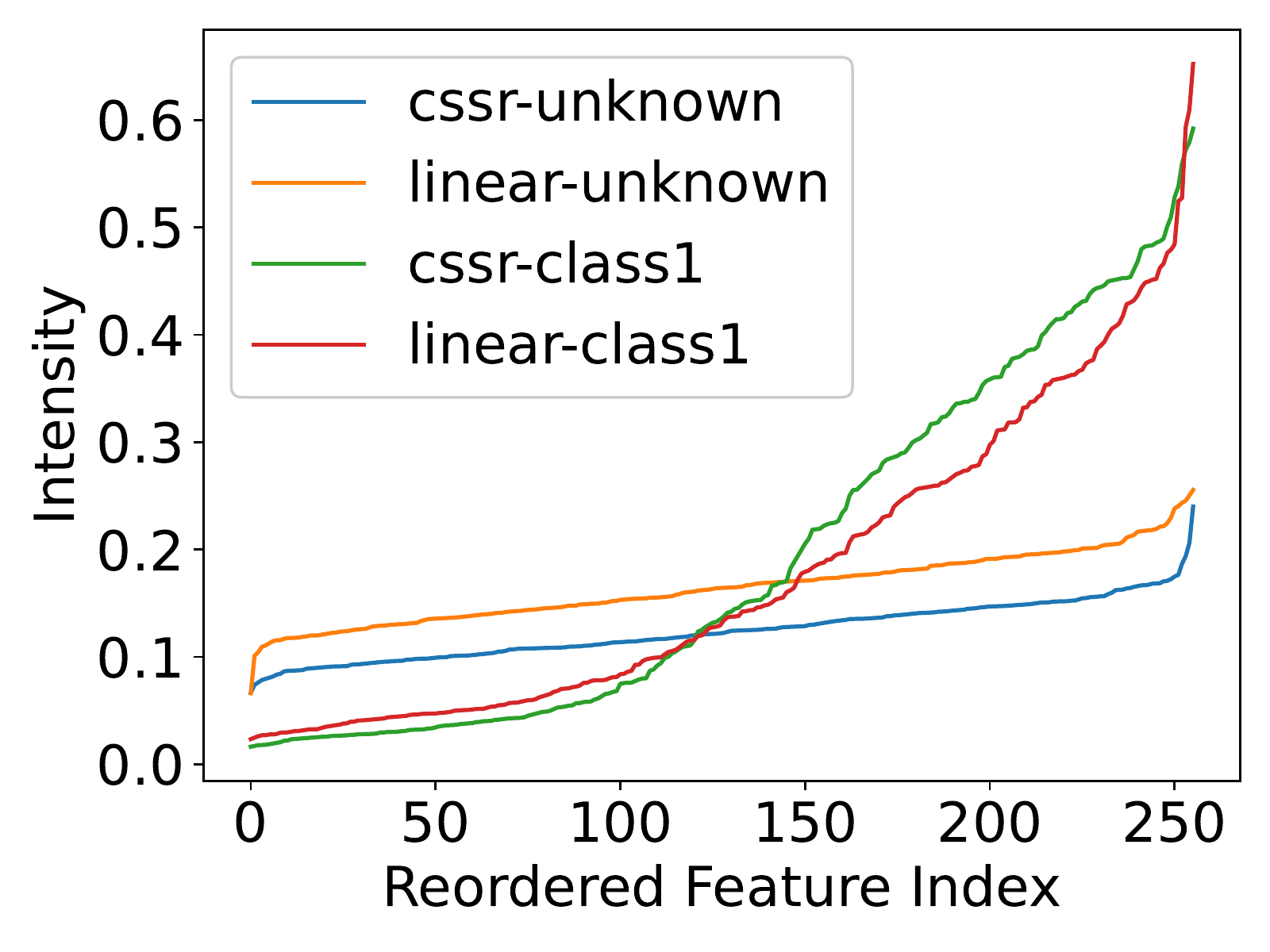}}
    \subfigure{\includegraphics[width=0.48\linewidth]{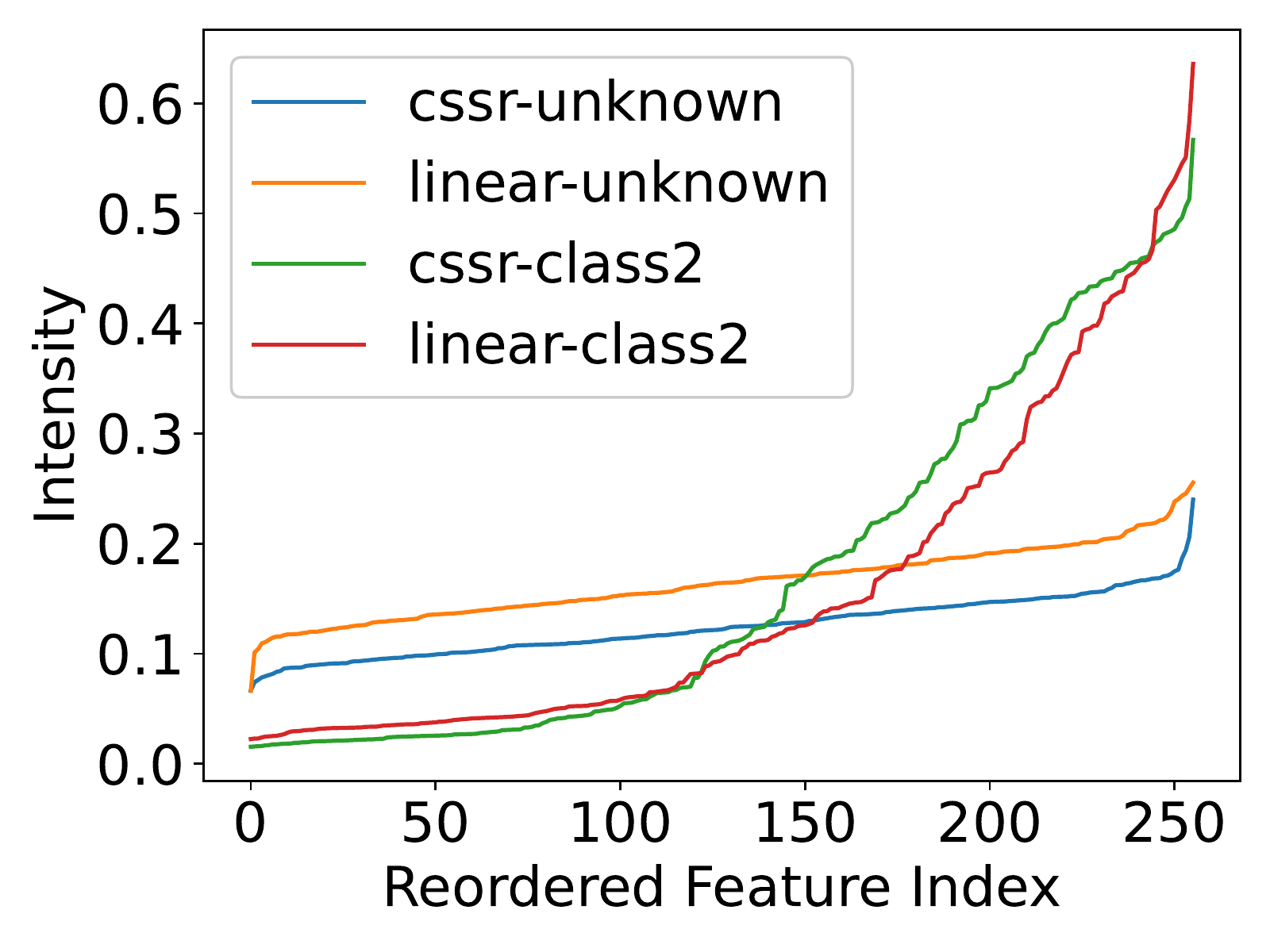}}
    \caption{Visualization of the class-specific feature activation for CSSR. We took six categories from Cifar10 as known classes and the remaining four as unknown. Note that the magnitude for each feature is normalized across the six known classes; we sorted each curve to have increasing order for better visualization.}
    \label{fig:csfeature}
\end{center}
\end{figure}

\subsubsection{Learning Class-Related Features}
\label{sec:learn_cs_features}
In the joint optimization of the backbone network and class-specific AEs, the learning process happens on both sides. In addition to the AEs being trained to fit known classes, the backbone network is trained simultaneously to bring the extracted features close to the class-specific manifold. This framework addresses the two problems mentioned in Section \ref{sec:introduction}, where diversity is required for well-learned representations. 
The diversity is naturally manifested in features located on the manifold surface. Meanwhile in the vertical direction of the manifold, the reconstruction error forces the representation to be compact. These properties make the semantic features class-related. Each category is described by and associated with a subset of global features, and a sample tends to only activate the associated features corresponding to its category.

We give an intuitive example here to explain how class-specific features are easier to learn. Consider a situation where the AEs have the simplest form: each encoder is an identity mapping of a class-specific subset of features, and the decoder identically maps back the features. Activating class-related features will not cause a reconstruction error, whereas activating class non-related features results in a reconstruction error. To reduce the reconstruction errors in this situation, the backbone learns to activate only class-related features. For joint optimization, the backbone network, as a much stronger fit, is used to reduce the AE's fitting complexity (\ie, making the AE as simple as possible). Finally, the model ends up extracting class-specific features. This property is also leveraged to detect unknown classes, which will be introduced in Section \ref{sec:detect_unknown}.

 \figref{fig:csfeature} shows the average activation intensity (\ie, the absolute activation value) for the known and unknown classes. Known classes are shown to be strongly activated on some specific features, while barely activated on the rest. Unknown classes, however, have low activation values on all features, as they are not trained and associated with any specific feature. Compared to the plain linear classification layer, the features extracted by CSSR are advantageous in several aspects: (1) class-related features for CSSR have a more uniform contribution, instead of focusing on a few features; (2) class-unrelated features for CSSR are less activated; and (3) the unknown class activation intensity of CSSR is clearly lower than that of a plain linear classifier, indicating a stronger association between known classes and learned semantic features.

Recently, a representation learning-based novelty detection method \cite{tack2020csi} also observed that well-learned representation makes the feature magnitude directly distinguishable between in-distribution and OOD data. Moreover, in the face of the problem of OOD data being available during training, Dhamija \etal \cite{dhamija2018reducing} proposed a loss to explicitly reduce the activation magnitude for OOD data. For CSSR, this property is implicitly satisfied without access to OOD data during training.

\begin{figure*}
    \centering
    \includegraphics[width=0.95\linewidth]{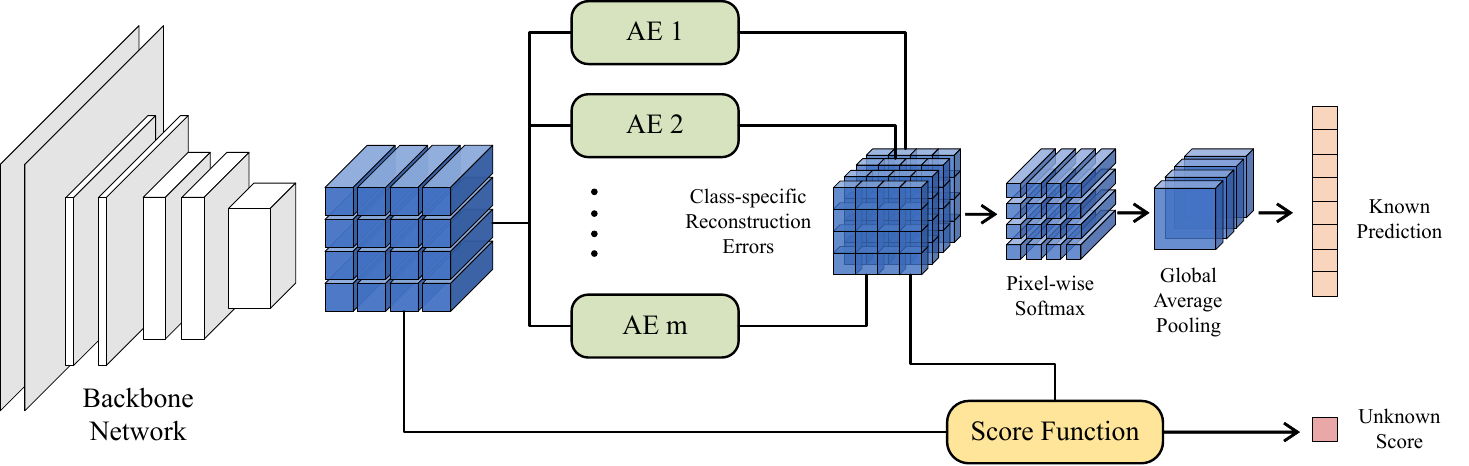}
    \caption{Overall architecture of our proposed model. The backbone network ($\mathcal{B}$) takes an image as input and extracts its semantic feature map $\mbfz$. The AE ($\mathcal{A}_c$) with respect to class $c$ encodes and reconstructs the pixelwise semantic representation $\mbfz$. Subsequently, we take the pixelwise reconstruction errors of class-specific AEs as logits and make pixelwise SoftMax on the logits multiplied by $\gamma$. Then, global average pooling is applied to reduce the pixelwise predictions to a general prediction for the whole image. For \textbf{unknown inference}, the model uses the pixelwise reconstruction errors corresponding to the predicted class and semantic features as input, and scores the unknownness for the image. Finally, a threshold is determined ensuring 95\% known samples are correctly accepted; samples are rejected if their unknown scores are below the threshold.
    }
    \label{fig:architecture}
\end{figure*}

\subsection{Overall Framework}
\figref{fig:architecture} illustrates the training and inference procedures for the proposed method. Our framework comprises two major modules: a backbone network $\mathcal{B}$ for learning latent representations and class-specific AEs $\{\mathcal{A}_i\}_{i=1}^m$ for classifying known and detecting unknown classes. 

To fully utilize the semantic feature map $Z=\{\mbfz_{ij}\}$ extracted from $\mathcal{B}$, we equally treat the latent representation for each pixel. Then, we make a global prediction by averaging the pixelwise predictions:
\begin{align}
    p(y=i|Z,\mathcal{A}) = \frac{1}{|Z|}\sum_{\mbfz \in Z}{p(y=i|\mbfz,\mathcal{A})}.
\end{align}
Evidently, $p(y=i|Z,\mathcal{A})$ sums to one, as $p(y=i|\mbfz,\mathcal{A})$ sums to one individually.
Finally, the model is trained by minimizing the negative log-probability of the true class $c$ via gradient descent as follows:
\begin{align}
    \mathcal{L}= -\log{p(y=c|Z,\mathcal{A})}.
\end{align}
As representation for each pixel focuses on a local region of the input image, such operations can be viewed as augmenting the raw input by soft cropping and ensemble prediction results of augmented images. In the test phase, it naturally applies the test augmentation technique and, thus, improves the performance. The above operations can be implemented by $1\times1$ convolution, pixelwise SoftMax, and global average pooling. Moreover, we implement the AEs by using a linear encoder and decoder with the $\tanh$ activation function and no bias for simplicity.

Based on the idea in \cite{chen2020learning}, the proposed method can also be extended to a reciprocal learning framework. The class-specific AEs can be used to estimate otherness in place of reciprocal point sets. The only modification is that we set a negative hyper-parameter $\gamma$. Then, similar to \refeq{approx_prototype}, assuming $d(\mbfz,\mathcal{A}_c)$ is maximized, the reconstruction error can be approximated by
\begin{align}
    d(\mbfz,\mathcal{A}_c)= \|\mbfz - \mathcal{A}_c(\mbfz)\|_1 \approx \max_{\mbfv\in V_c} \|\mbfz - \mbfv \|_1. \label{eq:approx_reciprocal}
\end{align}
This process is equivalent to \refeq{reciprocal}. We refer to the reciprocal version of CSSR as RCSSR in the rest of this paper. 

\subsection{Detecting Unknown Classes}
\label{sec:detect_unknown}
Suppose that the semantic feature $Z$ and predicted label $c$ from a test sample are given. We construct score functions for unknown detection from two different perspectives: (1) reconstruction error and (2) class-specific feature statistics.

\subsubsection{Reconstruction Error-based Score Function}

One natural idea is to utilize the reconstruction error to detect unknown classes. As mentioned in Section \ref{sec:learn_cs_features}, unknown samples lead to inactivated semantic features. We observe that poorly activated semantic features cause low reconstruction errors, resulting in failure to detect unknowns. 
As we implement AEs by using a linear decoder and a linear encoder activated by $\tanh$, the approximate linear relationship between $\|\mbfz\|_1$ and $\|\mbfz - \mathcal{A}_c(\mbfz)\|_1$ is considered. 
The encoder $f$ and decoder $g$ satisfy $f(\lambda \mbfz)=\lambda f(\mbfz),g(\lambda \mbfz)=\lambda g(\mbfz)$ as they are linear functions; for the $\tanh$ activation, $\tanh{(\lambda \mbfz)} \approx \lambda \tanh{(\mbfz)}$ assuming $f(\mbfz)$ is located around zero. Consequently,
\begin{align}
\|\lambda\mbfz - \mathcal{A}_c(\lambda \mbfz)\|_1 \approx \lambda\|\mbfz - \mathcal{A}_c(\mbfz)\|_1 \notag,
\end{align}
and we remove the scaling factor $\lambda$ by dividing $\|\mbfz\|_1$. The requirement for $\mbfz$ to be located around zero can be easily satisfied for unknown samples. Furthermore, we consider the unknown detection ability of the feature magnitude by multiplying an extra $\|\mbfz\|_1$ term.

Specifically, we define the first score function as follows. For CSSR, known classes should have low relative reconstruction error, while having high feature magnitude. This is given by
\begin{align}
    s_{p1}(\mbfz,c) =  -\frac{d(\mbfz,\mathcal{A}_c)}{\|\mbfz\|_1^2}.
\end{align}
Meanwhile for RCSSR, known class samples should have high relative reconstruction error and high feature magnitude:
\begin{align}
    s_{r1}(\mbfz,c) = \frac{d(\mbfz,\mathcal{A}_c)}{\|\mbfz\|_1} \times \|\mbfz\|_1 = d(\mbfz,\mathcal{A}_c).
\end{align}
Similar to the closed set classification process, fully utilizing the pixelwise features, we score feature map $Z$ pixelwise and individual scores are summed as the final score for the entire image, \ie, $s_{*}(Z,c)=\frac{1}{|Z|}\sum_{\mbfz \in Z}s_{*}(\mbfz,c)$, where $s_{*}$ represents either $s_{p1}$ or $s_{r1}$ (Note that we use $*$ represents selecting CSSR or RCSSR).

Compared with original prototype learning, $s_{p1}$ is different owing to its representation learning ability. However, $s_{r1}$ has the same form as \refeq{reciprocal} in reciprocal learning with the approximation in (\ref{eq:approx_reciprocal}) and prediction function $c=\arg\max_i d(\mbfz,\mathcal{A}_i)$.

\subsubsection{Activation Pattern-based Score Function}

Although the property of CSSR learning class-related features has been utilized by taking the feature magnitude previously, we consider a more delicate model on the class-specific activation patterns by considering the first- and second-order statistics. Low-order statistics have been applied in OOD detection \cite{quintanilha2018detecting}. However, the statistical information serves as input features for an OOD classifier, which requires OOD samples for training. Conversely, we collect statistical information to directly formulate the score function.

Suppose the semantic feature map $Z_i$ and predicted class $c_i$ are obtained by sample $\mbfx_i$ from the training set. As we are only concerned with the activation intensity for the features, we pre-process all feature maps by calculating the absolute values of their elements; in the following discussion, we assume the feature map have been pre-processed. Because class-specific patterns are required, the feature maps are grouped into different sets according to their predicted classes, \ie, feature sets $\mathcal{Z}^c=\{Z_i|c_i=c,i=1,2,...,n\},c=1,...,m$.

For the first-order statistics, we first take the class-specific mean activation intensity:
\begin{align}
    \mu_i = \sum_{Z\in \mathcal{Z}^i}\sum_{\mbfz \in Z}\frac{1}{|\mathcal{Z}^i||Z|}\mbfz.
\end{align}
To consider the different scales of the activation intensities for different features, a normalization cross categories is further applied:
\begin{align}
     \tilde{\mu}_i = \frac{\mu_i}{\sum_j \mu_j},
\end{align}
where the vector division is conducted elementwise. To detect unknownness, a sample activating what a known class activates is more likely to be the same known class. Specifically, the activation intensity of each feature is weighed by $\tilde{\mu}_c$, and the unknownness is evaluated by the weighted average intensity across features. Pixel-wise score integration is performed as well. Formally, the score function is defined by
\begin{align}
    s_2(Z,c) = \sum_{\mbfz \in Z}\frac{1}{|Z|}\mbfz^\top \tilde{\mu}_c. \label{eq:s2}
\end{align}

For the second-order statistics, motivated by Sastry and Oore \cite{sastry2020detecting}, we utilize Gram matrices to model inter-feature co-occurrence. Let $F\in \mathbb{R}^{D\times |Z|}$ be the feature intensity matrix with pixelwise vectors in feature map $Z$ concatenated on columns, and $D$ be the feature dimension. The Gram matrix for the $i$th sample is defined by $G = FF^\top$. The elements in a Gram matrix $G$ describe how much the corresponding two features (indexed by row and column) are likely to be simultaneously activated. We average the Gram matrices for the feature maps class-specifically, as a template for feature co-occurrence patterns. Then, for a test sample, we calculate its Gram matrix, and score its unknownness by the sum of elementwise multiplication using the pre-calculated template with respect to its predicted label. The process can be formulated as 
\begin{align}
    G^c &= \frac{1}{|\mathcal{Z}^c|}\sum_{Z\in \mathcal{Z}^c}G(Z),\\
    s_3(Z,c)&= Sum(G^c\odot G(Z)),  \label{eq:s3}
\end{align}
where $Sum(\cdot)$ is a function summing up matrix elements, and $\odot$ is the matrix elementwise multiplication. We observe that the above operation is equivalent to extending the pixelwise feature $\mbfz$ to a second-order polynomial space, \ie, $\mbfz\mbfz^\top=[z_iz_j]_{D\times D}$. The Gram matrix can be written as a summation of pixelwise extended features $G=\sum_{\mbfz \in Z}\mbfz\mbfz^\top$; thus, $G^c$ represents the first-order statistics for the extended feature space. Further, the score function can be seen as pixelwise scoring and integration by $ s_3(Z,c) = \sum_{\mbfz \in Z} Sum(G^c\odot \mbfz\mbfz^\top)$. In addition to the primary definition of Gram matrix, the extended higher-order Gram matrix proposed in \cite{sastry2020detecting} is optional here, \ie,
\begin{align}
    G=\left(F^p{F^p}^\top \right)^{\frac{1}{p}},
\end{align}
where the power on matrix is calculated elementwise. We found that the higher-order Gram matrix marginally improves the performance; we set $p=8$ by default in the rest of this paper.

\subsubsection{Integrated Score Function}
Before integrating the above scores, we perform a normalization for individual scores, so that the scales for different score functions are unified. Specifically, for each score function $s_*$, we obtain scores for \emph{random augmented} training samples to reduce the effect of overconfidence on well-fitted training samples. Then, the mean and standard deviation can be calculated and applied for normalization. The process can be expressed as 
\begin{align}
    \tilde{s}_*(Z,c) = \frac{s_*(Z,c)-E(s_*)}{Std(s_*)},
\end{align}
where $E(s_*)$ and $Std(s_*)$, respectively, represent the pre-computed mean and standard deviation with respect to $s_*$.
We now integrate all three score functions via a linear combination to obtain our final score function:
\begin{align}
    s_{all}(Z,c) = w_1\times \tilde{s}_{*1}(Z,c) + w_2 \times \tilde{s}_2(Z,c) + w_3 \times \tilde{s}_3(Z,c), \label{eq:s_all}
\end{align}
where $s_{*1}$ represents either $s_{p1}$ or $s_{r1}$, and $w_1,w_2,$ and $w_3$ are weights for each detection score. 

\begin{figure}
\begin{center}
    \includegraphics[width=0.95\linewidth]{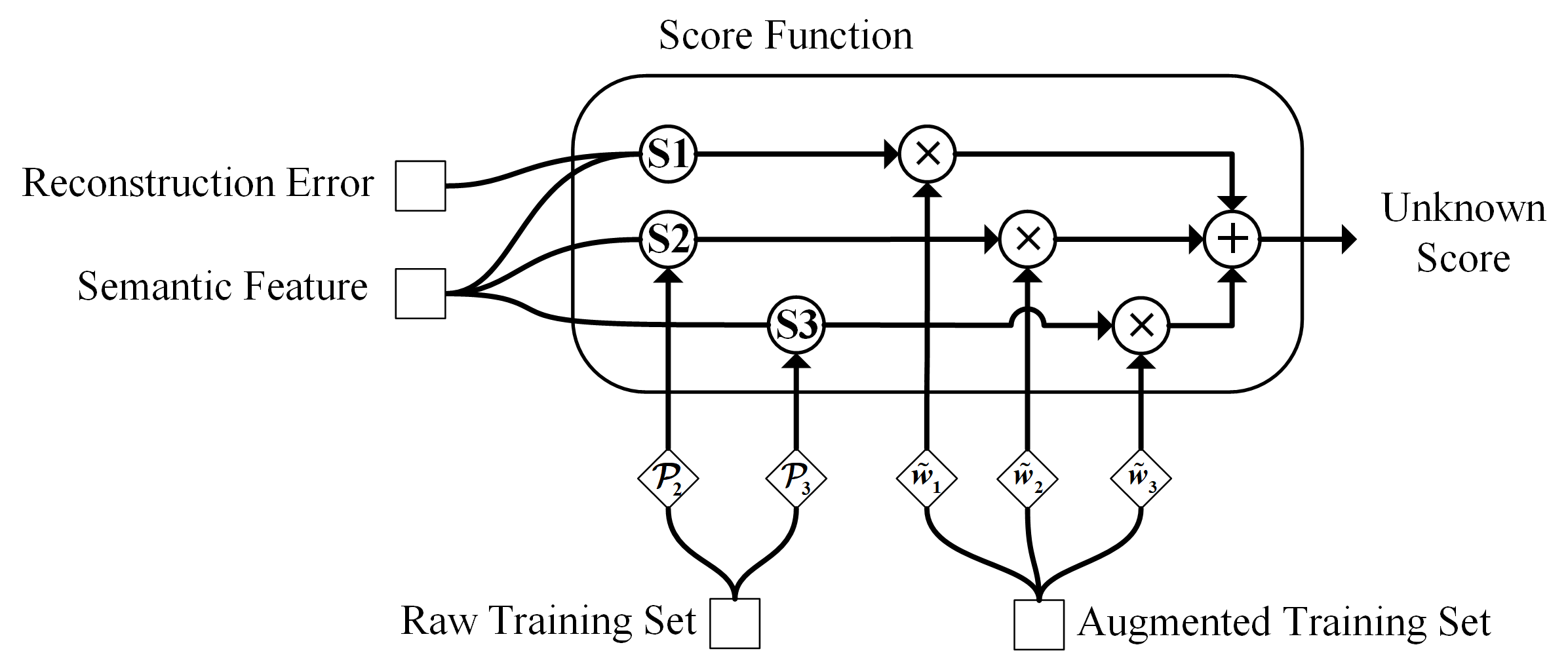}
    \caption{Unknown inference process. Before inference on test samples, we obtain first- and second-order statistics (corresponding to P2 and P3 in the figure, respectively) from the raw training set. 
    \todo{Then, the normalization coefficients ($\frac{1}{Std(s_*)}$) are obtained from the augmented training set for individual scores. The final score weights are calculated by the product of normalization coefficients and predefined weights, \ie, $\tilde{w}_*=\frac{w_*}{Std(s_*)}$, which forms the score function by $s_{all}(Z,c) = \sum_i \tilde{w}_i \times s_i(Z,c) $.}
    }
    \label{fig:scorefunc}
\end{center}
\end{figure}

The process is illustrated in \figref{fig:scorefunc}. We summarize our procedure for unknown detection as follows: (1) Run over the training set \emph{without data augmentation} to collect first-order and second-order statistics on semantic features. This step using non-augmented training samples preserves complete information for known classes. (2) Run over the training set \emph{with data augmentation} to collect the normalization parameters for individual scores. This step using augmented training samples reduces the effect of overconfidence. (3) Make inference using the integrated score in \refeq{s_all}. To reject unknown class samples, we take a threshold that guarantees 95\% known samples being accepted.

\subsection{Connection to Existing Methods}

\subsubsection{Connection to AE-based Methods}
AE-based open set recognition methods \cite{oza2019c2ae,oza2019deep,sun2020conditional} learn the classification task and reconstruction task simultaneously in the model. Recent works achieve this goal either by multi-task learning or multi-stage learning. In addition to the devouring problem caused by AE, as mentioned in Section \ref{sec:openspacerisk}, we analyze existing methods from the perspective of training frameworks.

Multi-task learning \cite{oza2019deep,sun2020conditional} jointly optimizes the classification task and reconstruction task by sharing common knowledge to learn a joint representation.
Consequently, these methods often perform slightly worse on close-set classification than plain classification.

The multi-stage approach \cite{oza2019c2ae} trains an encoder on a classification task, and then, trains the decoder, while freezing the parameters of the encoder. Although this strategy can maintain the close-set performance, it may have trouble in training the decoder, because the encoder is trained to extract discriminative classification features, which may not contain sufficient information to reconstruct the details.

As stated in Section \ref{sec:introduction}, separating the close-set classification and image-reconstruction task limits the performance as the encoded background information harms the classification task, and the trade-off between the two tasks is unavoidable. The proposed method avoids the above problems in two ways:
(1) CSSR reconstructs semantic features, rather than the raw image, which avoids reconstructing unnecessary background pixels and focuses on representative class-related semantic features;
(2) CSSR cascades the reconstruction and classification tasks, where the reconstruction task is trained basing on classification loss. Consequently, there is a positive correlation, rather than a negative correlation owing to the trade-off, between the performance of the two tasks.

\subsubsection{Connection to Representation Learning Methods}
\label{sec:connection_replearning}
Chen \etal \cite{chen2021exploring} recently proposed a new representation learning framework called SimSiam. We observed that CSSR and SimSiam share a similar architecture. SimSiam takes two randomly augmented views as input and processes them through a shared backbone network. Then, a prediction MLP head transforms the output of one view, and matches it with the other. The learning objective is to minimize the negative cosine similarity of the two output vectors. The prediction MLP head in SimSiam is a bottleneck structure (as recommended by the authors), which is similar to an AE. By ensuring that the two views are identical, SimSiam learns to maximize the cosine similarity between the raw semantic feature and reconstructed semantic feature. For each AE in CSSR, the reconstruction errors are minimized (similar to positive pairs) for samples of certain categories and maximized (similar to negative pairs) for samples of the rest. It learns class-specific representation, which is a subspace of the shared representation extracted by the backbone network. This explains how CSSR is enhanced to learn class-related semantic features, and makes feature intensity work directly to distinguish unknown classes.

\section{Experiments}

\begin{table}
  \setlength\tabcolsep{3pt}
  \centering
  \caption{AUROC comparison between different methods on unknown detection tasks. The best performance values are highlighted in bold.}
    \begin{tabular}{lccccc}
    \toprule
     Methods & SVHN  & CIFAR10 & CIFAR+10 & CIFAR+50 & TinyImageNet \\
    \midrule
     CROSR \cite{yoshihashi2019classification} & {89.9} & {88.3} & {91.2} & {90.5} & {58.9} \\
           C2AE \cite{oza2019c2ae} & {92.2} & {89.5} & {95.5} & {93.7} & {74.8} \\
           MLOSR \cite{oza2019deep}& {95.5} & {84.5} & {89.5} & {87.7} & {71.8} \\
           CGDL \cite{sun2020conditional} & {93.5} & {90.3} & {95.9} & {95.0} & {76.2} \\
           GFROSR \cite{perera2020generative} & {93.5} & {83.1} & {91.5} & {91.3} & {64.7}\\
    \midrule
     GCPL \cite{prototype} & {92.6} & {82.8} & -     & -     & - \\
           ARPL \cite{chen2021adversarial} & {96.7} & {91.0} & \textbf{97.1} & {95.1} & {78.2} \\
    \midrule
    
           Plain Softmax & 88.6 & 67.7 & 81.6 & 80.5 & 57.7 \\
           OSRCI \cite{neal2018open} & {91.0} & {69.9} & {83.8} & {82.7} & {58.6} \\
           PROSER \cite{zhou2021learning} & {94.3} & {89.1} & {96.0} & {95.3} & {69.3} \\
    \midrule
     CSSR & \textbf{97.9} & 91.3 & 96.3 & 96.2 & \textbf{82.3} \\
      RCSSR & 97.8 & \textbf{91.5} & 96.0 & \textbf{96.3} & 81.9 \\
    \bottomrule
    \end{tabular}%
  \label{tab:unknowndet}%
\end{table}%

\subsection{Implementation Details}

As CSSR modifies only the classification layer, various backbone networks can alternatively be used in implementing CSSR. Following Chen \etal \cite{chen2020learning}, we chose to train small-scale datasets with a Wide-ResNet \cite{zagoruyko2016wide} whose depth, width, and dropout rate we set to 40, 4, and 0, respectively, \ie, WRN40-4. However, for larger-scale datasets (\eg, TinyImageNet), we substituted the backbone with ResNet18 \cite{he2016deep} for efficiency. In the training phase, the stochastic gradient descent optimizer was used with momentum = 0.9. The model was trained for 200 epochs with batch size fixed to 128. The learning rate was set to 0.4 initially, and then, dropped by a factor of 10 at 130 and 190 epoch. We set $|\gamma|=0.1$ for all experiments. The AEs were implemented with linear encoders and linear decoders. To make AEs' embedding space $H$ bounded, we used $\tanh$ as the activation function. The dimension of the embedding space for AEs was set to 64 for the ResNet18 and WRN40-4 architecture. The score integration weights were set to be $1$s equally.
Previous methods used data augmentation techniques to improve open set discrimination. Following the settings of previous work~\cite{perera2020generative,zhou2021learning}, we apply a simple data augmentation technique in \cite{cubuk2020randaugment}.
\todo{A subset of transforms used in RandAugment is considered, which is: Brightness, Color, Equalize, Rotate, Sharpness, Shear and Contrast. For each input image, up to two of the transforms are sampled and applied to the image. }

In addition to prototype CSSR, RCSSR was also implemented and evaluated for a comprehensive comparison.

\subsection{Comparison with State-of-the-art Results}

\subsubsection{Unknown Detection}
\label{sec:sota_ud}
The evaluation protocol defined in \cite{neal2018open} was employed. Five image datasets were used in this experiment: SVHN \cite{netzer2011reading}, TinyImageNet \cite{pouransari2014tiny}, CIFAR10 \cite{krizhevsky09learning}, CIFAR+10, and CIFAR+50. For SVHN and CIFAR10, six classes were randomly sampled as the known classes, and the remaining four classes were used as the unknown classes. For TinyImageNet, 20 classes were sampled as the known classes, and the remaining 180 classes as the unknown classes. For the CIFAR+$M$ datasets, the model was trained on four non-animal classes from CIFAR10 as known classes, whereas $M$ animal classes from the CIFAR100 dataset \cite{krizhevsky09learning} were randomly selected as unknown classes. A threshold-independent metric, the area under the receiver operating characteristic (AUROC) curve, was used as the evaluation metric. It was calculated by plotting the true positive rate against the false positive rate by varying thresholds. The AUROC value is "1" if the knowns and unknowns are completely separable. Following \cite{neal2018open}, we averaged the results over five randomized trials.

We compared the frameworks related to our method, \ie, AE-based methods \cite{yoshihashi2019classification,oza2019c2ae,oza2019deep,sun2020conditional,perera2020generative} and prototype-like methods \cite{prototype,chen2021adversarial}, as well as two recent methods \cite{neal2018open,zhou2021learning}, using different architectures. The results are reported in Table \ref{tab:unknowndet}; the values other than CSSR are obtained from \cite{chen2021adversarial,prototype,oza2019deep,zhou2021learning}. Except for being slightly behind ARPL on CIFAR+10, CSSR outperforms all other approaches in the five datasets, especially on SVHN ($+1.2\%$), CIFAR+50 ($+1.0\%$), and TinyImageNet ($+4.1\%$).

\begin{table}
  \centering
  \caption{Open set classification results on the CIFAR-10 dataset with various unknown datasets added in the test phase.}
    \begin{tabular}{lcccc}
    \toprule
    Method & \multicolumn{1}{l}{IMGN-C} & \multicolumn{1}{l}{IMGN-R} & \multicolumn{1}{l}{LSUN-C} & \multicolumn{1}{l}{LSUN-R} \\
    \midrule
    Plain Softmax & 63.9 & 65.3 & 64.2 & 64.7 \\
    CROSR\cite{yoshihashi2019classification} & 72.1  & 73.5  & 72.0    & 74.9 \\
    GFROSR\cite{perera2020generative} & 75.7  & 79.2  & 75.1  & 80.5 \\
    C2AE\cite{oza2019c2ae}  & 83.7  & 82.6  & 78.3  & 80.1 \\
    CGDL\cite{sun2020conditional}  & 84.0    & 83.2  & 80.6  & 81.2 \\
    PROSER\cite{zhou2021learning} & 84.9  & 82.4  & 86.7  & 85.6 \\
    \midrule
    CSSR & 92.9  & 90.9  & \textbf{94.1}  & 93.5 \\
    RCSSR & \textbf{93.3}  & \textbf{91.5}  & 94.0    & \textbf{94.0} \\
    \bottomrule
    \end{tabular}%
  \label{tab:osr_comp}%
\end{table}%

\begin{table*}
  \centering
  \caption{\todo{Distinguishing CIFAR10 from near OOD dataset CIFAR100 and far OOD dataset SVHN under various metrics. }}
    \begin{tabular}{lcccccccc}
    \toprule
    \multirow{2}[4]{*}{Method} & \multicolumn{4}{c}{In:CIFAR10 / Out:CIFAR100} & \multicolumn{4}{c}{In:CIFAR10 / Out:SVHN} \\
\cmidrule(r){2-5} \cmidrule(r){6-9}          & DTACC & AUROC & AUIN  & AUOUT & DTACC & AUROC & AUIN  & AUOUT \\
    \midrule
    SoftMax & 79.8  & 86.3  & 88.4  & 82.5  & 86.4  & 90.6  & 88.3  & 93.6 \\
    GCPL\cite{prototype}  & 80.2  & 86.4  & 86.6  & 84.1  & 86.1  & 91.3  & 86.6  & 94.8 \\
    RPL\cite{chen2020learning}   & 80.6  & 87.1  & 88.8  & 83.8  & 87.1  & 92.0    & 89.6  & 95.1 \\
    ARPL\cite{chen2021adversarial}  & 83.4  & 90.3  & 91.1  & 88.4  & 91.6  & 96.6  & 94.8  & 98.0 \\
    CSI\cite{tack2020csi} & 84.4 & 91.6 & 92.5 & 90.0 & 92.8 & 97.9 & 96.2 & 99.0 \\
    OpenGAN\cite{kong2021opengan} & 84.2 & 89.7 & 87.7 & 89.6 & 92.1 & 95.9 & 93.4 & 97.1\\
    \midrule
    CSSR & 83.8  & 92.1    & 89.4  & 89.3  & \textbf{95.7} & \textbf{99.1} & 98.2  & \textbf{99.6} \\
    RCSSR & \textbf{85.3} & \textbf{92.3} & \textbf{92.9} & \textbf{91.0} & \textbf{95.7} & \textbf{99.1} & \textbf{98.3} & \textbf{99.6} \\
    \bottomrule
    \end{tabular}%
  \label{tab:comp_prototype_oodd}%
\end{table*}%

\subsubsection{Open Set Recognition}

In addition to detecting unknown classes, open set recognition requires a joint classification of known classes, while rejecting the unknowns. We followed the experimental set-up devised by Yoshihashi \etal \cite{yoshihashi2019classification}, where the models were trained on the entire CIFAR10 as known classes. In the test phase, the samples from other datasets were used as unknowns, \ie, ImageNet \cite{russakovsky2015imagenet} and LSUN \cite{yu2015lsun}. The two datasets were further cropped or resized to ensure that they had the same image size as the known samples; 10,000 samples (to maintain the consistency with the CIFAR10 test set) were selected forming ImageNet-Crop (IMGN-C), ImageNet-Resize (IMGN-R), LSUN-Crop (LSUN-C), and LSUN-Resize (LSUN-R). For a fair comparison, we used the version of the four datasets released by Liang \etal \cite{liang2018enhancing}. 
The performance was evaluated by macro-averaged F1-scores in 11 classes (including 10 known classes and the 1 unknown). The results are presented in Table \ref{tab:osr_comp}. The values other than CSSR are taken from \cite{zhou2021learning,oza2019deep,sun2020conditional}. 
It can observe that CSSR models outperformed existing methods by a large margin ($8.3\%$ on average). 

\subsubsection{OOD Detection}

In this section, we followed the experimental settings of Chen \etal \cite{chen2021adversarial} to compare with methods in an OOD detection setting. \todo{We also compared CSI \cite{tack2020csi} and OpenGAN~\cite{kong2021opengan} in this experiment. CSI shares a similar idea of utilizing well learned image representations for detecting OOD samples. To keep the comparison fair, we disabled the test augmentation technique for CSI during evaluation. For OpenGAN, we used the backbone trained in our baseline model as its feature encoder, and directly used corresponding test unknown dataset as its OOD validation set for model selection.} We considered two challenging pairs of OOD detection benchmarks \cite{hendrycks2016a}, including three common datasets: CIFAR10, CIFAR100, and SVHN. The models were trained on CIFAR10, whereas CIFAR100 and SVHN served as the near OOD and far OOD datasets, respectively, during the test phase. Note that the overlapping categories were removed from CIFAR100. In addition to AUROC, we used several other evaluation metrics, following Chen \etal \cite{chen2021adversarial}:
\begin{itemize}
    \item \textbf{Detection accuracy (DTACC).} This metric represents the maximum known/unknown classification accuracy over all possible thresholds. In calculating accuracy, the positive and negative samples were assumed to have equal probability to appear in the test set.
    \item \textbf{Area under the precision-recall curve (AUPR)}. The curve plot precision, $TP/(TP+FP)$, against recall, $TP/(TP+FN)$, with a varying threshold, where $TP,FP$, and $FN$ denote true positive, false positive, and false negative, respectively. AUPR is further calculated as AUIN and AUOUT, where in- and out-distribution samples are set as positive, respectively.
\end{itemize}

As presented in Table \ref{tab:comp_prototype_oodd}, we compared the results with those from ARPL \cite{chen2021adversarial}. \todo{The state-of-the-art methods all achieved similar performance on two OOD datasets.} For the near OOD dataset, CSSR performs comparable to APRL, and is significantly better than primary prototype point-based open set recognition (GCPL), while RCSSR outperforms APRL by an increment of 2.3\%. For the far OOD dataset, we observed that CSSR and RCSSR have similar performance; both of them outperformed traditional prototype-like methods by a large margin. 

\subsection{Ablation Study}

The contributions from different components and score functions of CSSR are analyzed in this section. We first compare various architectures for the model.

\textbf{Datasets:} We trained the models on CIFAR10. For experiments on CIFAR10, we used all 10 classes in CIFAR10 as known classes, and then, tested on SVHN, LSUN-Resize, ImageNet-Resize, LSUN-Fix (LSUN-F), and ImageNet-Fix (IMGN-F). LSUN-Fix/ImageNet-Fix contains randomly sampled and resized images from LSUN/ImageNet produced by Tack \etal \cite{tack2020csi}, and the two datasets are more challenging than the original version released by Liang \etal \cite{liang2018enhancing}. %


\textbf{Ablation Terms.} 
(1) \emph{Classification Layer:} We compared traditional classification models with plain linear classification layers as baselines, and we kept the backbone and hyperparameters fixed for a fair comparison.
(2) \emph{Classification Strategy:} We used the proposed pixelwise prediction strategy (pixelwise SoftMax; then average pooling, namely SM-AP) or plain prediction strategy (average pooling; then SoftMax, namely AP-SM). The pixelwise prediction strategy affects both training and testing.
(3) \emph{Reconstruction Error Measurement:} We measured reconstruction errors with MSE or MAE (by default) for CSSR.

\begin{table*}[htbp]
  \centering
  \caption{Ablation study on various architectures. The first row specifies the datasets used as unknown classes; "Close Acc" represents closed set test accuracy ($\%$). For unknown detection, we provide AUROC values for evaluation.}
  \begin{tabular}{lccccccc}
    \toprule
    Method & Close Acc & SVHN   & LSUN-R & IMGN-R & LSUN-F & IMGN-F & Average \\
    \midrule
    Linear  & 96.77 & 97.0  & 95.3 & 94.2 & 92.3 & 93.2 & 94.4\\
    Linear SM-AP & 96.96 & 96.8  & 95.7 & 94.8 & 92.9 & 93.4 & 94.7\\
    \midrule
    CSSR AP-SM & 96.69 & 98.9   & 98.5  & 97.1  & 89.7  & 89.5 & 94.7\\
    CSSR MSE & 96.85 & 98.9   & 98.5  & 97.3  & 95.4  & 94.4 & 96.9\\
    CSSR & 96.86 & 99.1   & 98.8  & 97.5  & 96.2    & 95.3 & \textbf{97.4}\\
    \midrule
    RCSSR AP-SM & 96.84 & 97.4   & 97.1  & 95.5  & 88.1 & 89.2 & 93.5 \\
    RCSSR MSE & 96.82 & 98.7    & 97.8  & 96.5  & 92.0   & 91.1 & 95.2 \\
    RCSSR & 97.02 & 99.1   & 99.1  & 98.1  & 96.0   &95.0 & 97.3 \\
    \bottomrule
    \end{tabular}%

  \label{tab:ablation_component}%
\end{table*}%

The results are shown in Table \ref{tab:ablation_component}. The table shows the following: (1) As a non-linear classification layer, CSSR slightly improves closed set performance. (2) Pixelwise classification slightly improves closed set classification performance, while largely improving the performance of unknown detection. (3) Using a distance measure of MAE generally outperforms MSE, demonstrating that MSE is a good choice for detecting unknown samples.

Next, the effect of different score functions is analyzed. We compared them by fixing trained CSSR and specifying different score functions for decision making. ImageNet30 (a subset of ImageNet introduced by Hendrycks \etal \cite{hendrycks2019using}) was utilized in this experiment, where 10 classes were sampled as known and the remaining 20 classes as unknown. The class split was kept the same for all experiments, where the top 10 classes in alphabetical order were selected as known classes for simplicity. To adapt to ImageNet-30, which has higher image resolution, RandomCrop was replaced by RandomResizedCrop, following standard data augmentations in training ImageNet. A plain ResNet18 with linear classification layer was considered as the baseline. Six different score functions were compared: relative reconstruction error (RRE, $\frac{d(\mbfz,\mathcal{A}_c)}{\|\mbfz\|_1}$), feature magnitude (FM, $\|\mbfz\|_1$), $s_{*1}$ ($s_{p1}$ for CSSR, $s_{r1}$ for RCSSR, and maximum SoftMax probability for baseline), $s_2$ (\refeq{s2}), $s_3$ (\refeq{s3}), and $s_{all}$ (\refeq{s_all}).

\begin{table}[htbp]
  \centering
  \caption{AUROC comparison on different score functions for ablation models trained on ImageNet-30. }
    \begin{tabular}{lcccccc}
    \toprule
    Methods & RRE   & FM    & $s_{*1}$  & $s_2$  & $s_3$  & $s_{all}$ \\
    \midrule
    Linear & -     & 39.3  & 93.5  & 92.5  & 87.1  & - \\
    \midrule
    CSSR AG-SM & 93.6  & 90.6  & 92.2  & 94.0    & 91.9  & 94.6 \\
    CSSR MSE & 85.1  & 81.9  & 92.2  & 95.1  & 93.9  & 94.7 \\
    CSSR & 91.3  & 91.5  & 94.8  & 95.5  & 94.6  & 95.3 \\
    \midrule
    RCSSR AG-SM & 84.5  & 84.7  & 95.1  & 92.4  & 92.1  & 94.7 \\
    RCSSR MSE & 89.7  & 81.3  & 95.0    & 94.2  & 93.7  & 94.6 \\
    RCSSR & 86.8  & 90.2  & 95.1  & 94.6  & 94.4  & 95.0 \\
    \bottomrule
    \end{tabular}%
  \label{tab:ablation_scorefunc}%
\end{table}%

The results are illustrated in Table \ref{tab:ablation_scorefunc}, where we observe the following: (1) The features learned by the plain linear classification layer are less class-related; the feature magnitude is not sensitive to detect unknown classes, and the representation-based score functions are less discriminative to the unknown classes. (2) CSSR, which improves representation learning ability, also improves the two representation-based score functions $s_2,s_3$. (3) For $s_{*1}$, integrating both RRE and FM significantly improves the open set performance for both CSSR and RCSSR. (4) Pixelwise prediction and MAE are good at improving the quality of the learned representations, and therefore improving the performance of representation-based score functions. 

\begin{figure}
\begin{center}
    \subfigure[CSSR]{\includegraphics[width=0.95\linewidth]{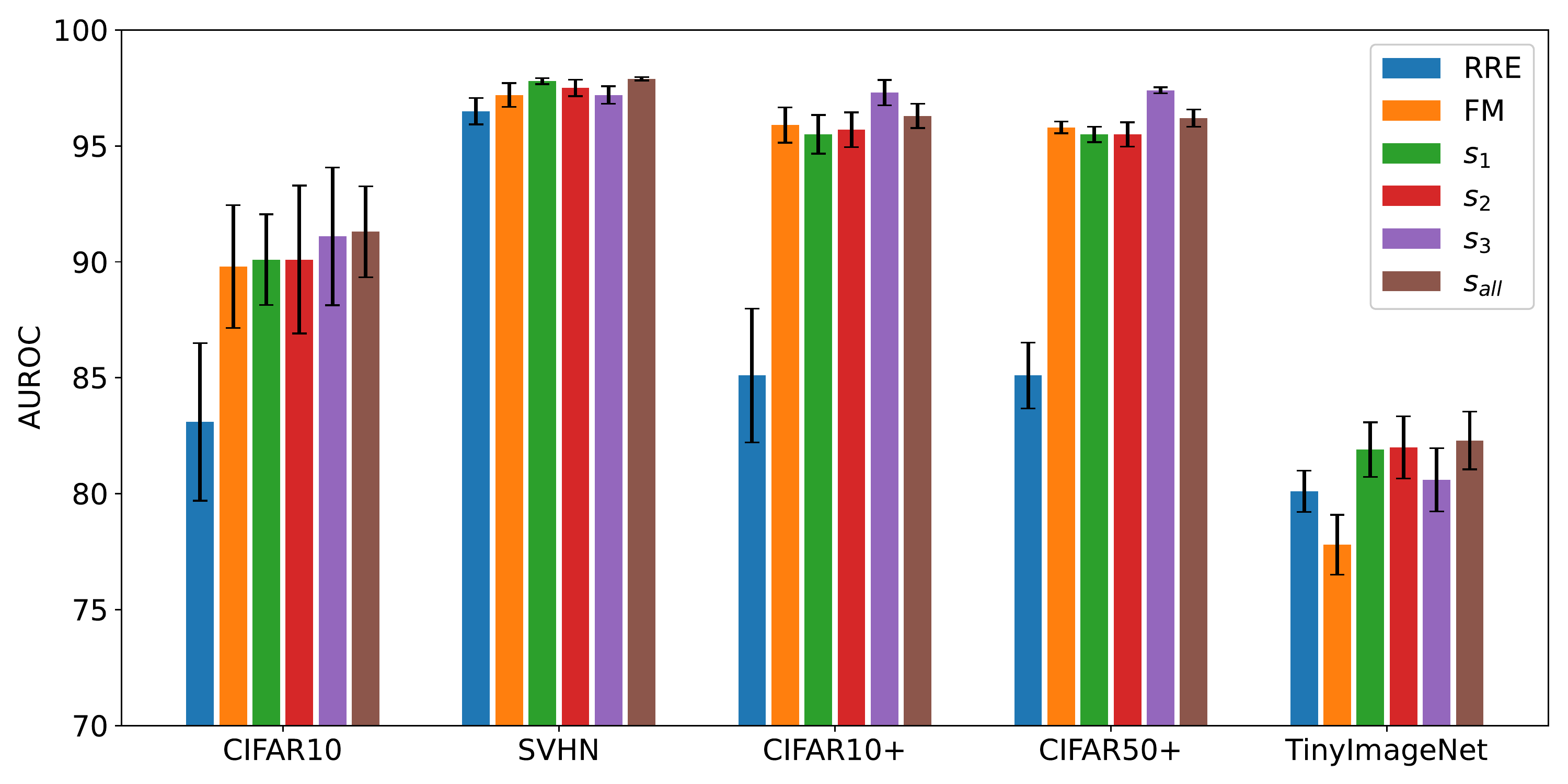}}
    \subfigure[RCSSR]{\includegraphics[width=0.95\linewidth]{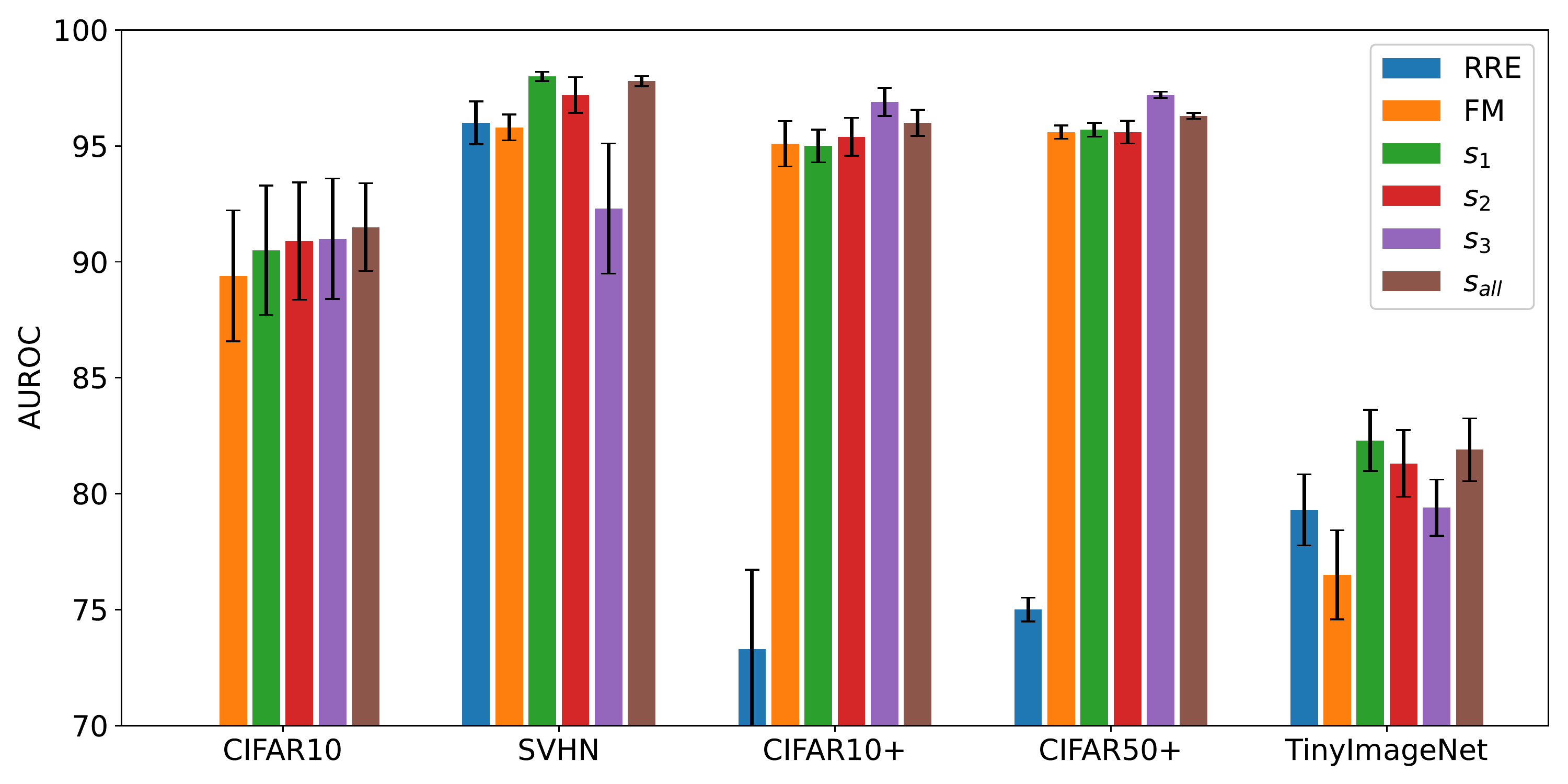}}
    \caption{The performance of individual score function in the unknown detection experiment. Note the standard deviations are calculated by the five randomized trials under different known-unknown splits.}
    \label{fig:score_ablation}
\end{center}
\end{figure}

Although the score fusion in \refeq{s_all} is not guaranteed to improve all of the individual scores, it approximately maintains the best individual score. We demonstrate that the three score functions have different performances on different datasets. Retaining the best performance of the three score functions improves the overall performance across datasets and reduces the variance. In \figref{fig:score_ablation}, we further show how different score functions perform in the unknown detection experiment (Table \ref{tab:unknowndet}) to demonstrate performance variation under different datasets. For example, $s_3$ gains an advantage in CIFAR+10 and CIFAR+50, but not in SVHN and TinyImageNet. The fused score, however, is at least the second best and has minimal standard deviation.

\subsection{Further Analysis}

\subsubsection{Closed Set Performance}

\todo{Maintaining closed set performance is equally important for open set recognition. Here, we compared the closed set performance on CIFAR10. We disabled extra data augmentation techniques to make the comparison fair. For the comparison methods, we took the reported results from their original paper. To consider the implementation difference, we also provided a baseline result under our implementation. As is shown in Table \ref{tab:close_performance}, CSSR shows superior performance on closed set classification. CSSR marginally improves performance by 0.2\% (CSSR) and 0.5\% (RCSSR), indicating better representation learning improves closed set prformance for CSSR.}

\begin{table}[htbp]
  \centering
  \caption{\todo{Closed performance comparison on CIFAR10.}}
    \begin{tabular}{llr}
    \toprule
    Method  & Accuracy \\
    \midrule
    CROSR \cite{yoshihashi2019classification} & 94.0 \\
    CGDL \cite{sun2020conditional}   & 91.2 \\
    GCPL \cite{prototype}  & 93.3 \\
    ARPL \cite{chen2021adversarial}  & 94.0 \\
    \midrule
    Our baseline & 95.1 \\
    CSSR  & 95.3 \\
    RCSSR & 95.6 \\
    \bottomrule
    \end{tabular}%
  \label{tab:close_performance}%
\end{table}%

\subsubsection{Performance Against Openness}

\begin{figure}
\begin{center}
    \includegraphics[width=0.90\linewidth]{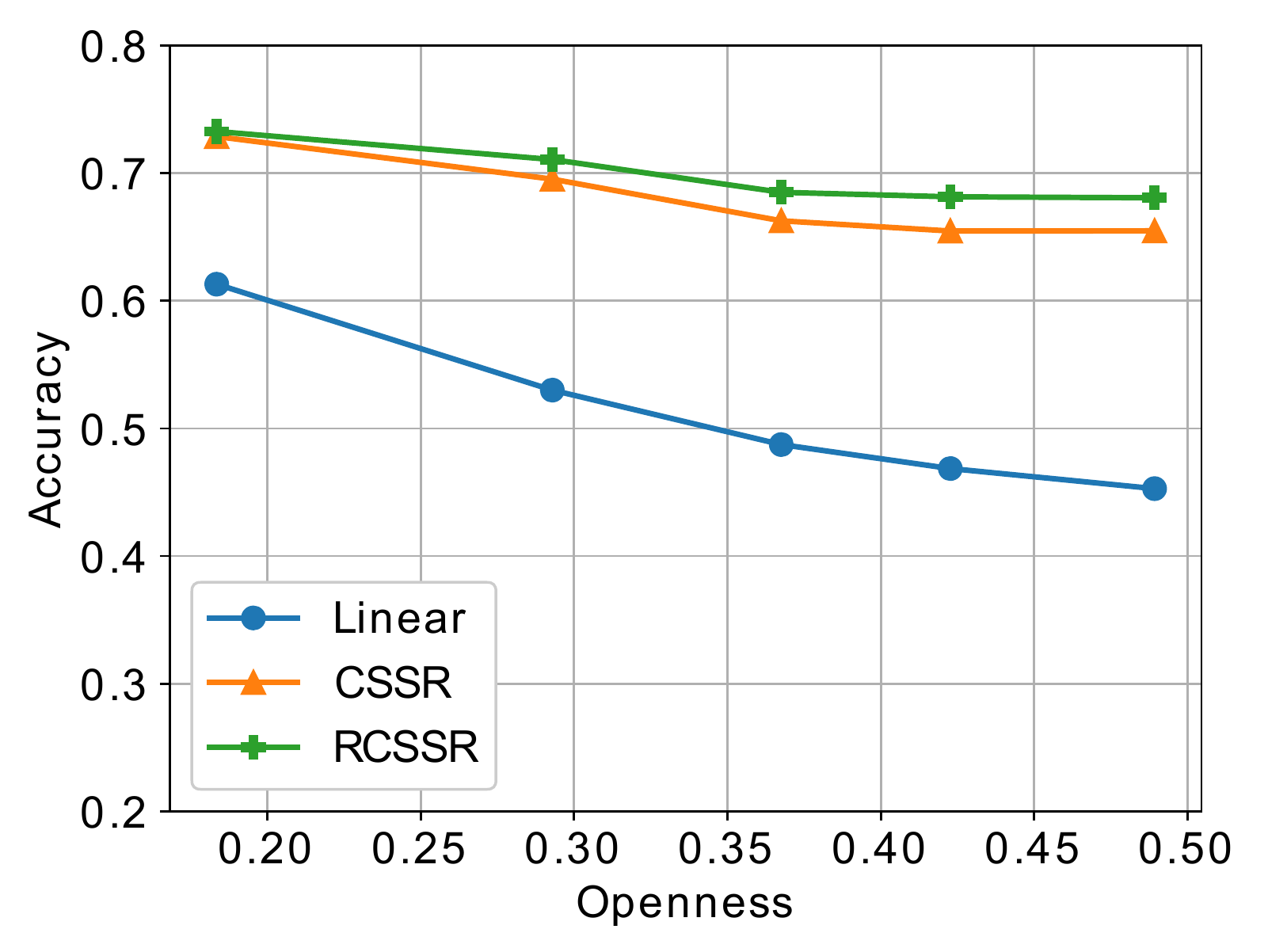}
    \caption{Open set recognition accuracy against varying Openness for CSSR and baseline.}
    \label{fig:openness}
\end{center}
\end{figure}

Openness \cite{scheirer2013toward}, as a measure representing the complexity of the open set task, is defined by
\begin{align}
    Openness=1-\sqrt{\frac{2*N_{train}}{N_{test}+N_{target}}},
\end{align}
where $N_{train}$ is the number of known classes seen during training, $N_{test}$ is the number of classes that will be observed during testing, and $N_{target}$ is the number of classes to be recognized during testing. Using common experimental settings \cite{sun2020conditional,zhou2021learning}, we conducted the experiment on CIFAR100, where 15 classes were randomly sampled as known classes. The number of unknown classes varied from 15 to 85, meaning that the openness varied from 18\% to 49\%. The recognition performances of 16 classes (15 known classes and 1 unknown) were evaluated by classification accuracy. The results are shown in \figref{fig:openness}. CSSR shows good performance with increasing openness, while the performance drops rapidly when using a plain linear classification layer.

\subsubsection{Performance on Large-scale Datasets}

To evaluate our model on a large-scale classification task, we conducted experiments on ImageNet-1000 \cite{russakovsky2015imagenet}. As a more challenging dataset, ImageNet-1000 includes 1,000 classes with more than one million training images and 50,000 validation images. \todo{We first followed the experiment settings used by Yang \etal \cite{prototype}, where the first 100 classes were selected as known, and the remaining 900 as unknown.} We adopted ResNet18 as the backbone network. To adapt to large-scale classification and save parameters, we reduced the dimension of the embedding space for AEs from 64 to 32. The learning rate still started at 0.4, but dropped by a factor of 10 at the 100 and 150 epoch for sufficient training. Considering the larger and more complex semantic space introduced by ImageNet-1000, we disabled the representation-based score functions, \ie, $w_2=w_3=0$. The remaining hyperparameters were kept the same as in the experiments on ImageNet30. In addition to considering the model plain linear classification layer as our baseline, we took the reported results for the original prototype-based method \cite{prototype} to make comparisons. Note that the implementation in \cite{prototype} leveraged ResNet50 as the backbone, which is much stronger than ResNet18 as used in our study. 

Two additional metrics are used to evaluate the models more comprehensively: (1) \textbf{TNR@TPR95} is the probability that an unknown sample is correctly rejected with the true positive rate (TPR) being $95\%$; (2) \textbf{open set classification rate (OSCR)}, as defined in \cite{dhamija2018reducing}, was adopted. We denote the score threshold by $\delta$. \textbf{Correct classification rate (CCR)} is the fraction of known samples that are correctly classified with unknown detecting scores above the given threshold $\delta$. \textbf{False positive rate (FPR)} is the fraction of unknown samples whose unknown detecting scores are greater than threshold $\delta$. The CCR and FPR values under different thresholds were reduced to one specific value by taking the area under CCR against the FPR curve.

\begin{table}[tbp]
  \centering
  \caption{Results on \todo{splited} ImageNet-1000. Performances for both unknown detection and open set recognition are evaluated.}
    \begin{tabular}{lcccc}
    \toprule
          & \multicolumn{2}{c}{Unknown detection} & \multicolumn{2}{c}{Open set recognition} \\
    \cmidrule{2-3}\cmidrule{4-5}
         Method & AUROC & TNR@TPR95 & Macro-F1 & OSCR \\
    \midrule
    SoftMax \cite{prototype} & 79.7  & -     & -     & - \\
    GCPL \cite{prototype}  & 82.3  & -     & -     & - \\
    \midrule
    Our baseline & 91.0    & 51.8  & 40.1  & 77.4 \\
    CSSR & \textbf{93.7}  & \textbf{62.4}  & \textbf{43.0}    & \textbf{78.1} \\
    RCSSR & 93.1  & 58.4  & 40.7  & 77.6 \\
    \bottomrule
    \end{tabular}%
  \label{tab:imgnet1000}%
\end{table}%

The results are shown in Table \ref{tab:imgnet1000}. We can first observe that our implementation for baseline clearly outperforms previous studies in the literature, indicating an underestimation of the open set performance on large-scale datasets. The proposed CSSR and RCSSR outperform the baseline in detecting unknown classes, especially CSSR. It is also observed that the improvement in open set recognition for RCSSR appears relatively small compared to the task of unknown detection. This is due to the degradation of the closed set performance for RCSSR. 

\todo{Further, we considered a large-scale classification setting, where samples from ImageNet-100 and iNaturalist \cite{Horn2018TheIS} are used as known and unknown classes, respectively. In addition to the baseline, we compared two state-of-the-art methods, ARPL \cite{chen2021adversarial} and OpenGAN \cite{kong2021opengan}. To achieve efficient training, we used a ResNet18 backbone pretrained on ImageNet-1000 for all methods. To train CSSR, APRL and the baseline, we fixed the pretrained backbone and fine-tuned the classification layer for 4 epochs. While for OpenGAN, the pretrained backbone served as the fixed feature encoder. Note that OpenGAN requires a validation set that servers as the known unknowns to choose appropriate discriminator for unknown inference. In this experiment, we simply took the best-performed checkpoint on the test set to avoid a choice of validation set. As shown in Table \ref{tab:imgnet1000all}, CSSR/RCSSR improves existing methods with a significant margin by 5.2\% on AUROC, 23.2\% on TNR@TPR95, and 2.9\% on DTACC, respectively.}

\begin{table}[tbp]
  \centering
  \caption{\todo{Open set recognition for large-scale classification, where samples from ImageNet-1000 are known and samples from iNaturalist serve as the unknowns.}}
    \begin{tabular}{lccc}
    \toprule
    Method & AUROC & TNR@TPR95 & DTACC \\
    \midrule
    SoftMax & 87.2  & 41.2  & 79.1 \\
    ARPL~\cite{chen2021adversarial}  & 88.8  & 50.6  & 80.2 \\
    OpenGAN~\cite{kong2021opengan} & 89.3  & 32.0    & 84.3 \\
    \midrule
    CSSR  & 93.8  & 70.7  & 86.5 \\
    RCSSR & \textbf{94.5}  & \textbf{73.8}  & \textbf{87.2} \\
    \bottomrule
    \end{tabular}%
  \label{tab:imgnet1000all}%
\end{table}%

\begin{table}[!htbp]
  \centering
  \caption{\todo{Entire training time for different methods on ImageNet-1000. The value $+x$h demonstrates that the method requires extra $x$ hours fine-tuning in addition to pre-training a plain model.}}
    \begin{tabular}{lll}
    \toprule
    Method & Training Strategy & Time \\
    \midrule
    Plain & From Scratch & 140h \\
    CSSR  & From Scratch & 225h \\
    \midrule
    ARPL  & Fix Backbone & +6h \\
    OpenGAN & Fix Backbone& +1h  \\
    CSSR  & Fix Backbone & +8h \\
    \bottomrule
    \end{tabular}%
  \label{tab:entire_time}%
\end{table}%

\begin{figure}[htbp]
\begin{center}
   
    \subfigure[]{\includegraphics[width=0.48\linewidth]{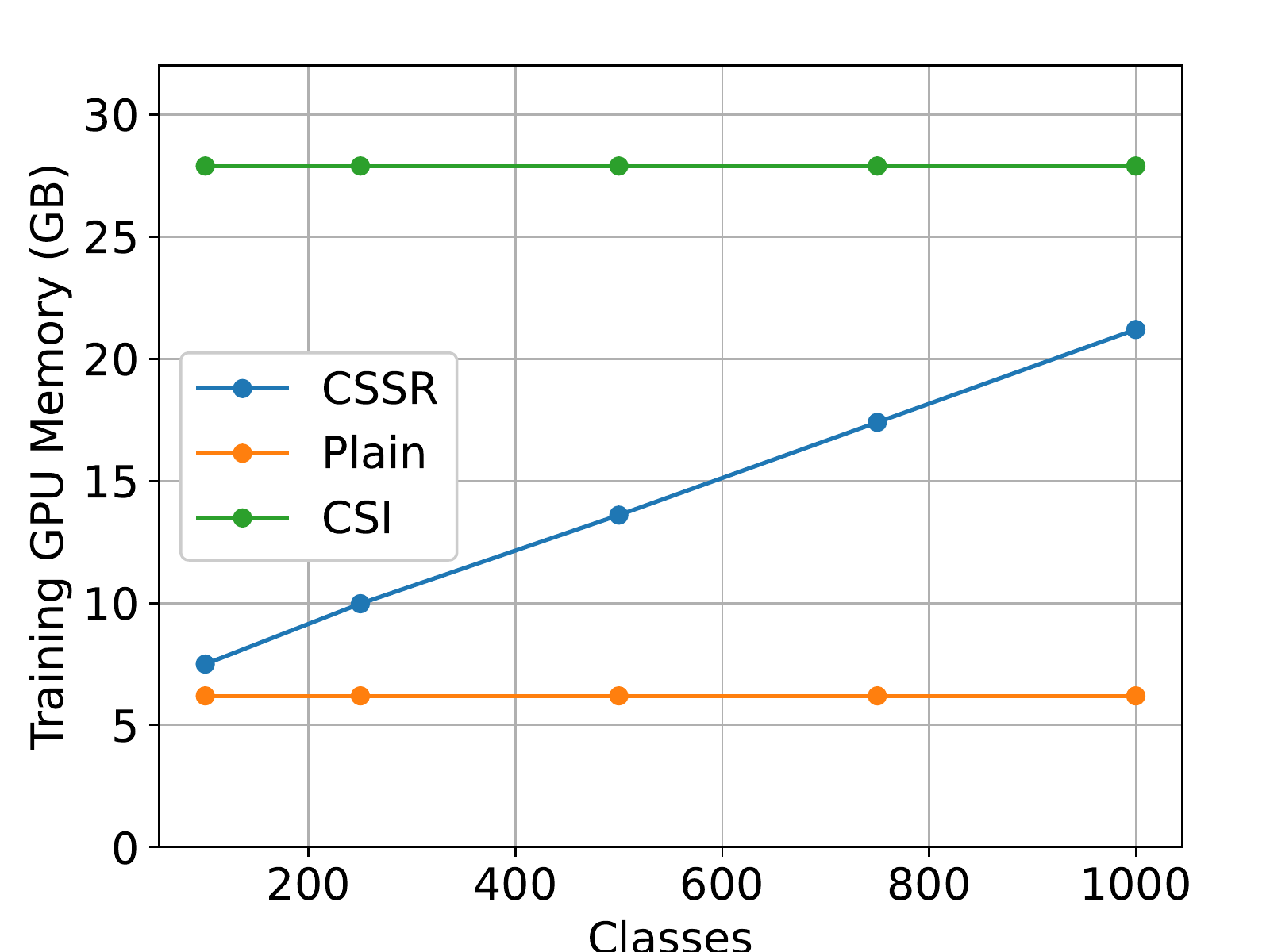}}
    \subfigure[]{\includegraphics[width=0.48\linewidth]{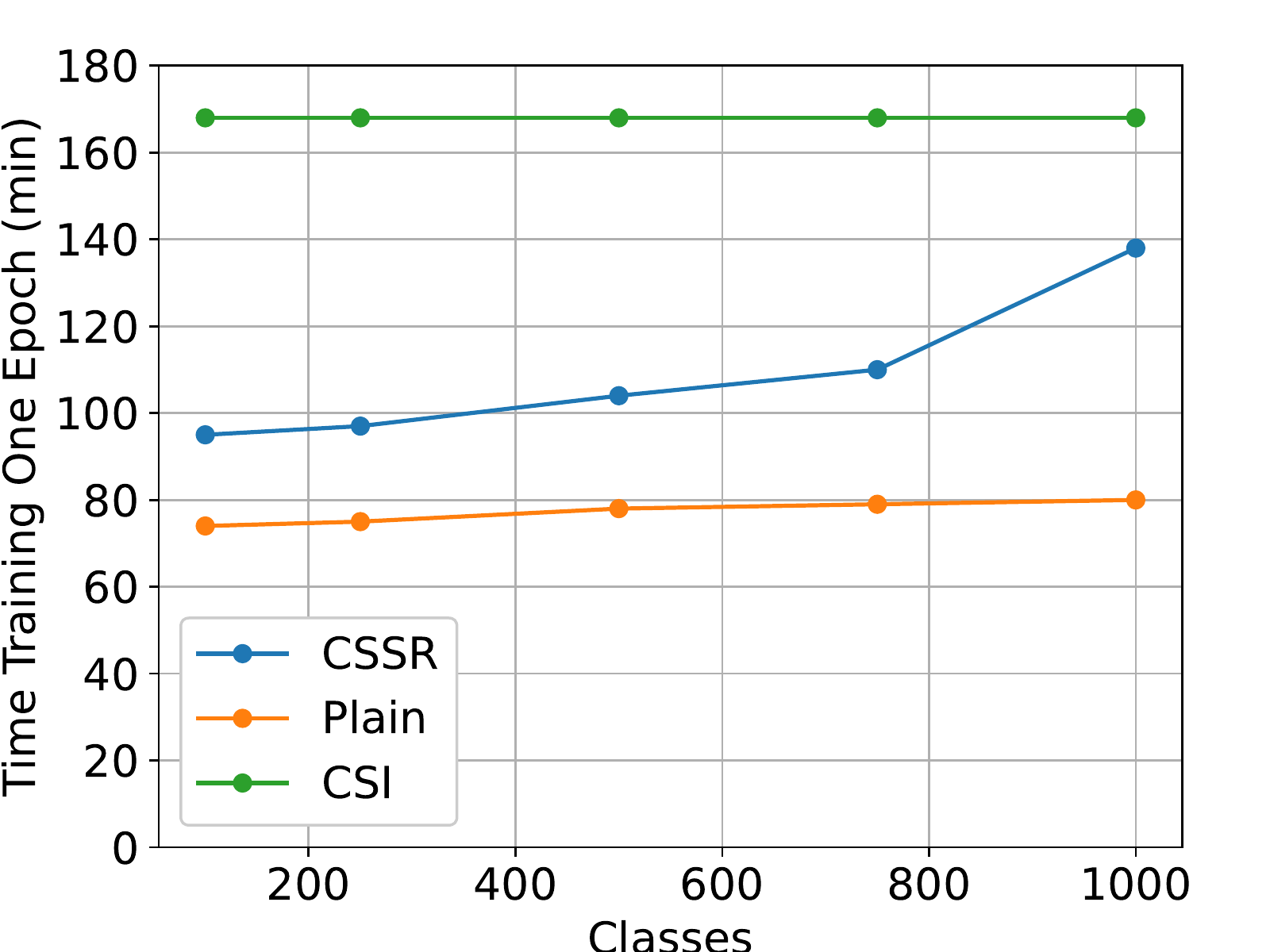}}
    \caption{Comparison of the amount of GPU memory used (a) and the amount of time consumed in one epoch training (b). The values are collected by training ResNet18-based models on ImageNet-1000 with batch size 128. The number of samples are kept the same when varying the number of classes.}
    \label{fig:memory_epochtime}
\end{center}
\end{figure}

\todo{One possible concern might be that the number of parameters of CSSR would increase linearly as the number of categories increases. In fact, the number of extra parameters for CSSR is only a half compared to CGDL~\cite{sun2020conditional}, a recent autoencoder-based method, in a 1000-way classification task (approximately 60M for CSSR and approximately 175M for CGDL). A major concern might be the increase in training time and GPU memory requirement for CSSR. As shown in Fig. \ref{fig:memory_epochtime}, we observed over 50\% increase in training time for CSSR compared to plain models. 
Notably, compared to some recent OOD methods, \eg,  CSI~\cite{tack2020csi} as shown in the figure, CSSR is still light-weight. To make CSSR practical for extreme large-scale classification, one may pretrain a plain classification model and then finetune on a CSSR classification head to reduce the resources consumption, as we have done on ImageNet-1000. With finetuning technique, CSSR shows little efficiency difference compared to other SOTA methods, as shown in TABLE~\ref{tab:entire_time}. A more promising way is to adopt a commonly existing class hierarchical that organizes classes from coarse to fine. CSSR can only perform on the coarse level with simple classifier on the fine level. We remain this to future work.}


\begin{figure*}[!htb]
\begin{center}
    \includegraphics[width=0.95\linewidth]{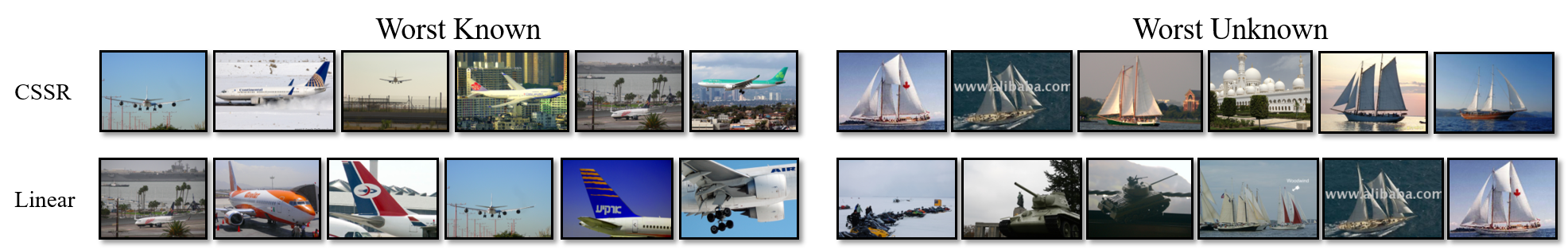}
    \caption{Known and unknown samples that CSSR or plain linear model fails to identify.
    }
    \label{fig:failures}
\end{center}
\end{figure*}

\subsubsection{Analysis of Failures}

To demonstrate the difference between CSSR and plain classification models with a linear classification layer, the models trained in the ImageNet30 experiment (Table \ref{tab:ablation_scorefunc}) were chosen. Then, for both CSSR and plain linear, we picked the worst recognized known and unknown samples for visualization. Specifically, we selected the six samples with the lowest detection scores in category flight (known during training), and six samples with the highest scores among unknown samples, representing the worst failures. We further constrained the unknown samples to be predicted as category flight for better comparison. \figref{fig:failures} shows the selected images. For the failures on the known samples, CSSR mainly focuses on distant images, where the target object is relatively small. However, the plain model also fails on occasions where the plane is too close. As the known classes are modeled with intra-diversity allowed for CSSR, visual variants have less influence on the overall recognition. For the failures on the unknown samples, the sailboats cause the most confusion, which might be because of the similar backgrounds and textures. However, in the plain model, the tanks are not expected to be confused, and such mistakes can lead to severe issues in real-world applications.

\section{Conclusion}

In this study, we integrated AE and prototype-learning frameworks to propose a novel end-to-end learned deep network, called CSSR, for open set recognition. CSSR specifies an individual AE to each known class as a substitute of the class-specific point set in the traditional prototype learning framework. The AEs are inserted on the top of the backbone DNN to reconstruct the learned semantic representations of the images. These class-specific AEs can be considered prototype learning with class-specific learnable AE manifold represented point sets. 
We also noted that the MSE distance, which is commonly used in prototype learning, potentially causes inconsistent distribution between prototype points and ground-truth data distribution. 
To address this problem, the MAE distance is used to replace MSE and can guarantee consistency between prototype points and data points. Furthermore, our framework can be modified to RCSSR with the idea of reciprocal point learning. As the proposed method can learn boost feature representations, various representation-based score functions were explored in this paper, including first-order and second-order statistics. 

The results of experiments conducted on multiple datasets demonstrate that the proposed method outperforms other state-of-the-art methods. Note that CSSR requires no extra loss other than the discriminative classification loss and acts as a classifier. Therefore, CSSR can further be extended easily using various classification techniques. 
In real-world applications, high-dimensional and large-scale images are two typical challenges for OSR.
We plan to focus on the problem of open set recognition with class hierarchical structures in future work. \todo{Moreover, our work lacks the concern of training open set classifier with available known outliers. We plan to tackle this by exploiting image backgrounds in future work.}


\bibliographystyle{IEEEtranS}
\bibliography{ref}

\vspace{-10mm}
\begin{IEEEbiography}[{\includegraphics[width=1in,height=1.25in,clip,keepaspectratio]{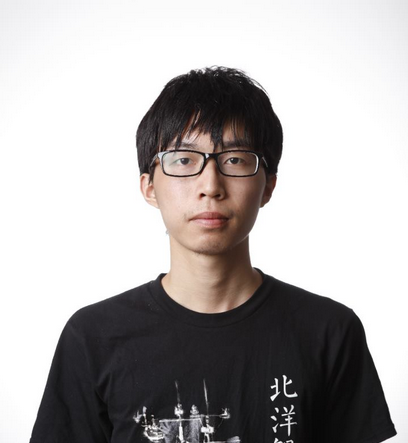}}]{Hongzhi Huang} received the B.S. degree in computer science and technology from Tianjin University in 2020, and is pursuing his M.S. degree in Tianjin University. His research interests are on open set recognition and out-of-distribution detection in computer vision.

\end{IEEEbiography}

\vspace{-10mm}

\begin{IEEEbiography}[{\includegraphics[width=1in,height=1.25in,clip,keepaspectratio]{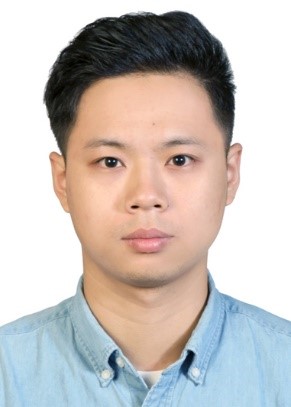}}]{Yu Wang} received the B.S. degree in communication engineering, the M.S. degree in software engineering, and Ph.D. degree in computer applications and techniques from Tianjin University in 2013 and 2016, and 2020, respectively. He is currently an assistant professor of Tianjin University. His research interests focus on hierarchical learning and large-scale classification in computer vision and industrial applications, data mining, and machine learning.

\end{IEEEbiography}
\vspace{-10mm}

\begin{IEEEbiography}[{\includegraphics[width=1in,height=1.25in,clip,keepaspectratio]{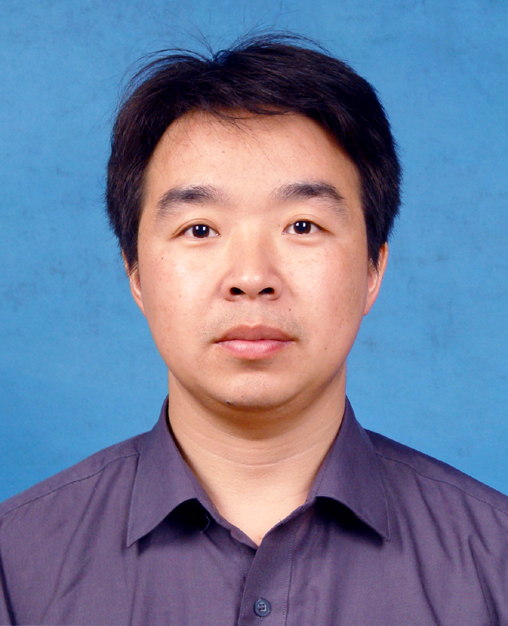}}]{Qinghua Hu} received the B.S., M.S., and Ph.D. degrees from the Harbin Institute of Technology, Harbin, China, in 1999, 2002, and 2008, respectively. He has published over 200 peer-reviewed papers. His current research is focused on uncertainty modeling in big data, machine learning with multi-modality data, intelligent unmanned systems. He is an Associate Editor of the IEEE TRANSACTIONS ON FUZZY SYSTEMS, Acta Automatica Sinica, and Energies.
\end{IEEEbiography}

\vspace{-10mm}
\begin{IEEEbiography}[{\includegraphics[width=1in,height=1.25in,clip,keepaspectratio]{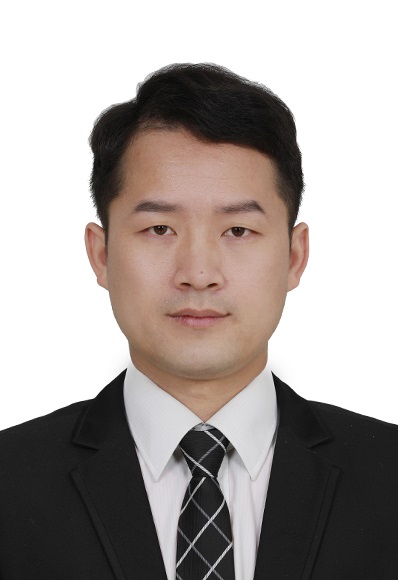}}]{Ming-Ming Cheng} received his PhD degree from Tsinghua University in 2012, and then worked with Prof. Philip Torr in Oxford for 2 years. He is now a professor at Nankai University, leading the Media Computing Lab. His research interests includes computer vision and computer graphics. He received awards including ACM China Rising Star Award, IBM Global SUR Award, \etc. He is a senior member of the IEEE and on the editorial boards of IEEE TPAMI and IEEE TIP.
\end{IEEEbiography}

\vfill

\end{document}